\documentclass[10pt,twocolumn,letterpaper]{article}

\usepackage[pagenumbers]{cvpr} % To force page numbers, e.g. for an arXiv version

\usepackage{multirow}
\usepackage{wrapfig}
\usepackage{dblfloatfix}
\usepackage{cuted}
\usepackage{xspace}
\usepackage{caption}
\usepackage{thm-restate}
\usepackage{footmisc}
\usepackage{amsmath}
\usepackage{amssymb}
\usepackage{amsthm}

\newcommand{\x}{\mathbf{x}}
\newcommand{\z}{\mathbf{z}}

\newcommand{\cc}{\mathbf{c}}
\newcommand{\F}{\mathbf{F}}

\newtheorem{lemma}{Lemma}

\definecolor{cvprblue}{rgb}{0.21,0.49,0.74}
\usepackage[pagebackref,breaklinks,colorlinks,allcolors=cvprblue]{hyperref}

\usepackage[capitalize,noabbrev]{cleveref}
\usepackage{algorithm}
\usepackage{algorithmic}
\usepackage{pifont}
\newcommand{\cmark}{\ding{51}}% 체크마크
\newcommand{\xmark}{\ding{55}}% X마크

\def\paperID{8041} % *** Enter the Paper ID here
\def\confName{CVPR}
\def\confYear{2026}

\title{ICM-SR: Image-Conditioned Manifold Regularization \\ for Image Super-Resolution}

\author{
Junoh Kang\footnotemark[1]~~$^1$ \quad Donghun Ryou\footnotemark[1]~~$^2$ \quad Bohyung Han$^{1,2}$ \\
Computer Vision Laboratory, $^1$ECE \& $^2$IPAI, Seoul National University
\\
{\tt\small \{junoh.kang, dhryou, bhhan\}@snu.ac.kr}
}

\begin{document}

\maketitle
\def\thefootnote{*}\footnotetext{indicates equal contribution.}\def\thefootnote{\arabic{footnote}}

% !TEX root = ../main.tex

\begin{abstract}

Real world image super-resolution~(Real-ISR) often leverages the powerful generative priors of text-to-image diffusion models by regularizing the output to lie on their learned manifold.
However, existing methods often overlook the importance of the regularizing manifold, typically defaulting to a text-conditioned manifold. 
This approach suffers from two key limitations. 
Conceptually, it is misaligned with the Real-ISR task, which is to generate high quality~(HQ) images directly tied to the low quality~(LQ) images.
Practically, the teacher model often reconstructs images with color distortions and blurred edges, indicating a flawed generative prior for this task.
To correct these flaws and ensure conceptual alignment, a more suitable manifold must incorporate information from the images.
While the most straightforward approach is to condition directly on the raw input images, their high information densities make the regularization process numerically unstable.
To resolve this, we propose image-conditioned manifold regularization~(ICM), a method that regularizes the output towards a manifold conditioned on the sparse yet essential structural information: a combination of colormap and Canny edges.
ICM provides a task-aligned and stable regularization signal, thereby avoiding the instability of dense-conditioning and enhancing the final super-resolution quality.
Our experiments confirm that the proposed regularization significantly enhances super-resolution performance, particularly in perceptual quality, demonstrating its effectiveness for real-world applications. 
We will release the source code of our work for reproducibility.

\end{abstract}
% !TEX root = ../main.tex

\section{Introduction}

\begin{figure}[t]
	\centering
    	\includegraphics[width=0.8\linewidth]{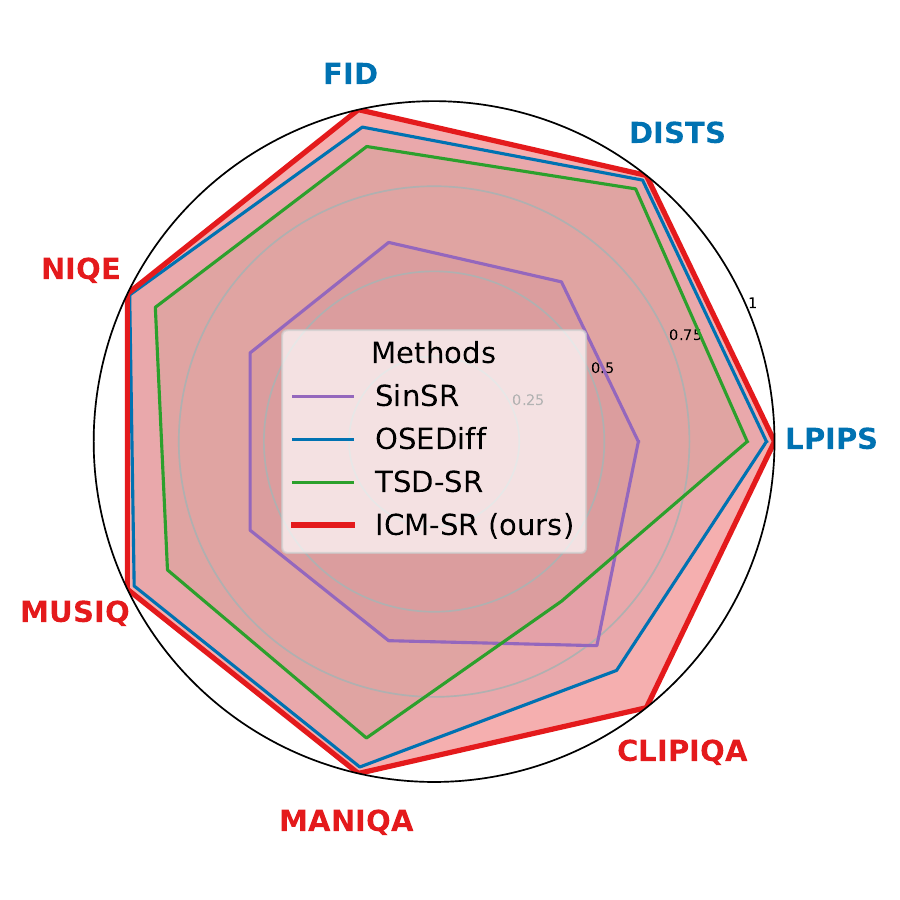} % 
	\vspace{-5mm}
    	\caption{
			Performance comparison on DRealSR benchmark~\cite{wei2020component}. 
			The red and blue metrics are no-reference and reference perceptual metrics, respectively.
			ICM-SR~(ours) stands out for perceptual metrics, highlighting its strong performance in practical scenarios.
    	}
	\vspace{-2mm}
    	\label{fig:raidar_side}
\end{figure}

\begin{figure*}[t]
\renewcommand{\arraystretch}{0}
\setlength{\tabcolsep}{0pt}

\scalebox{0.95}{
\begin{tabular}{@{}l@{\hspace{1mm}}c@{\hspace{0.5mm}}c@{\hspace{0.5mm}}c@{\hspace{0.5mm}}c@{\hspace{0.5mm}}c@{\hspace{3mm}}c@{\hspace{0.5mm}}c@{\hspace{0.5mm}}c@{\hspace{0.5mm}}c@{\hspace{0.5mm}}c@{\hspace{0.5mm}}c@{}}

    && HQ & Colormap & Canny & & &HQ & Colormap & Canny & & \\
    &&\includegraphics[width=0.10\linewidth]{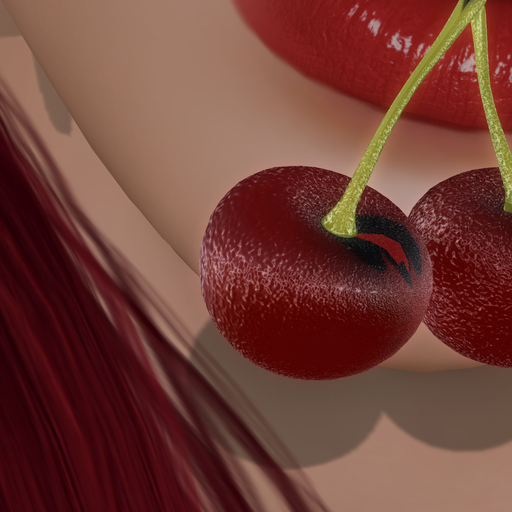} &
    \includegraphics[width=0.10\linewidth]{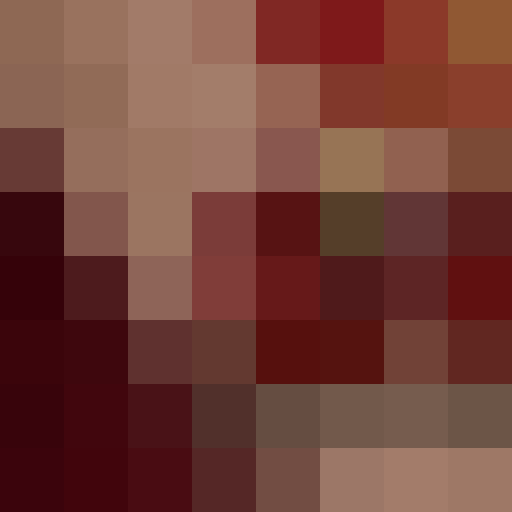} &
    \includegraphics[width=0.10\linewidth]{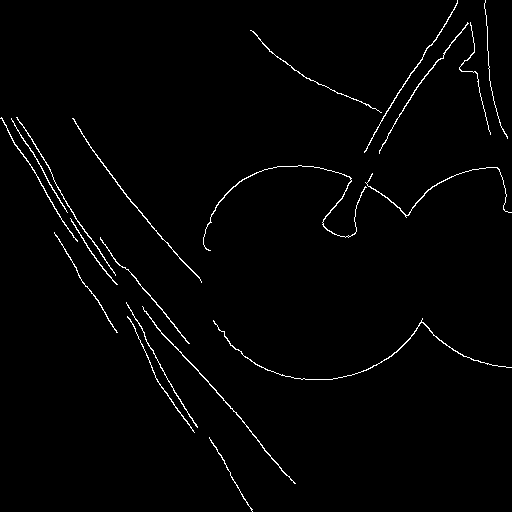} & & & 
    \includegraphics[width=0.10\linewidth]{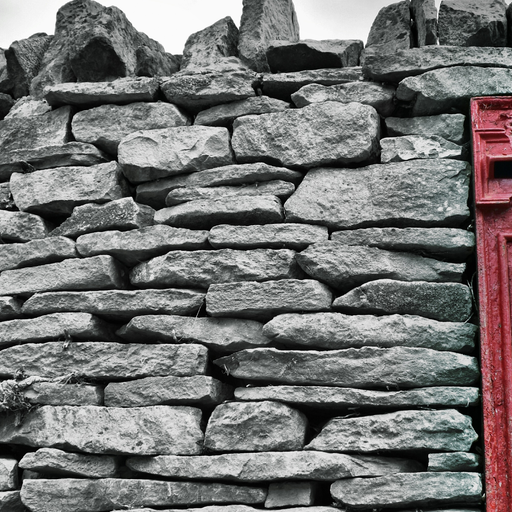} &
    \includegraphics[width=0.10\linewidth]{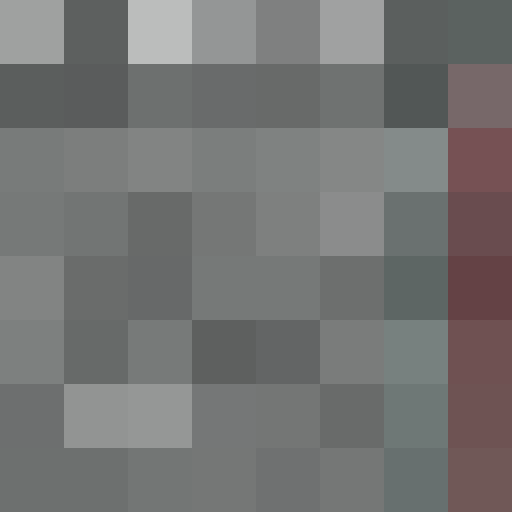} &
    \includegraphics[width=0.10\linewidth]{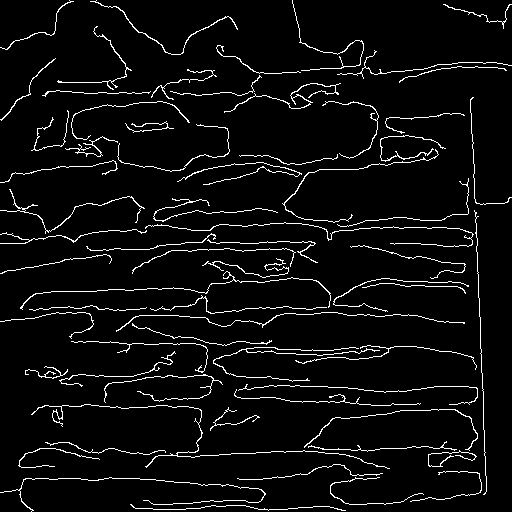} & & \\[2mm]
    
    & $t=500$ & $t=620$ & $t=740$ & $t=860$ & $t=980$ & $t=500$ & $t=620$ & $t=740$ & $t=860$ & $t=980$ \\[1mm]

    \rotatebox{90}{\hspace{2mm} \footnotesize Text-cond.} &
    \includegraphics[width=0.10\linewidth]{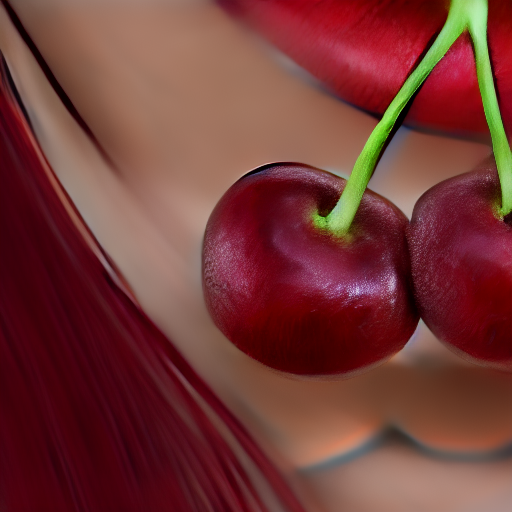} &
    \includegraphics[width=0.10\linewidth]{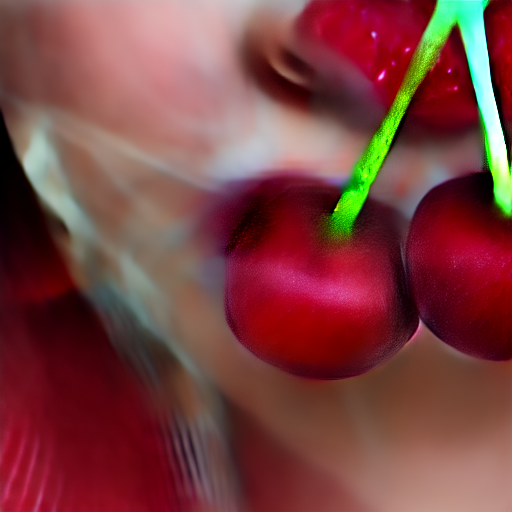} &
    \includegraphics[width=0.10\linewidth]{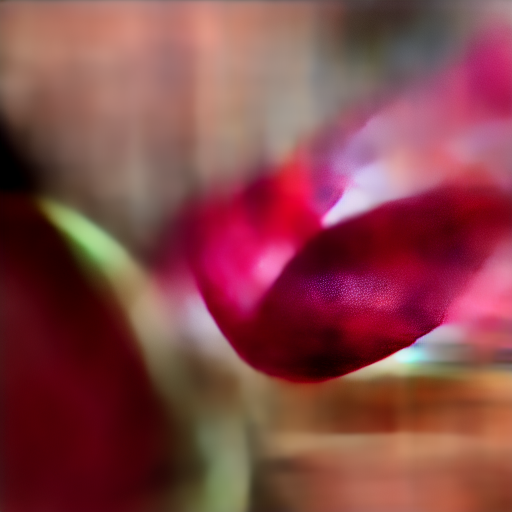} &
    \includegraphics[width=0.10\linewidth]{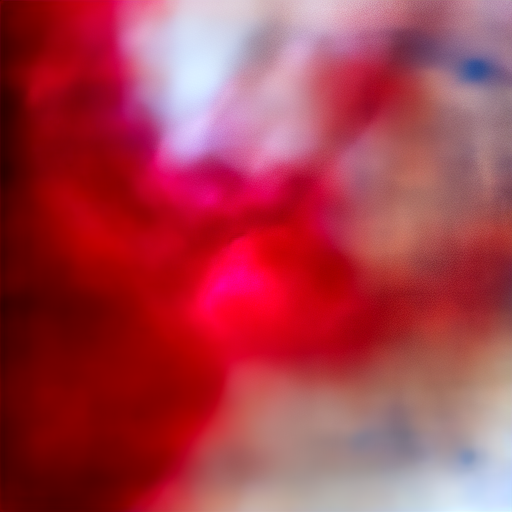} &
    \includegraphics[width=0.10\linewidth]{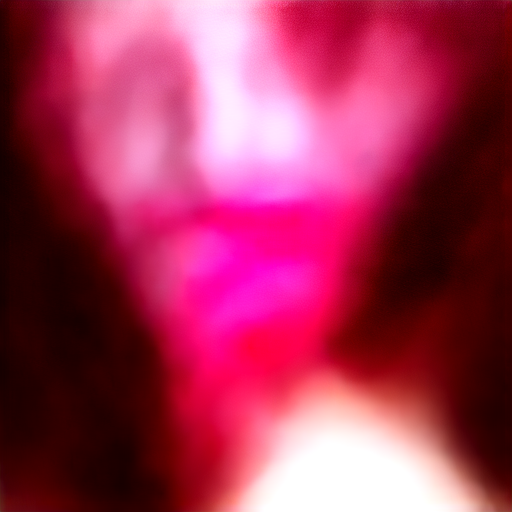} &
    \includegraphics[width=0.10\linewidth]{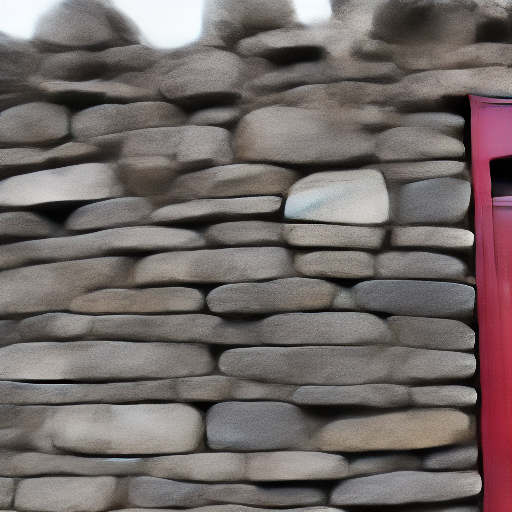} &
    \includegraphics[width=0.10\linewidth]{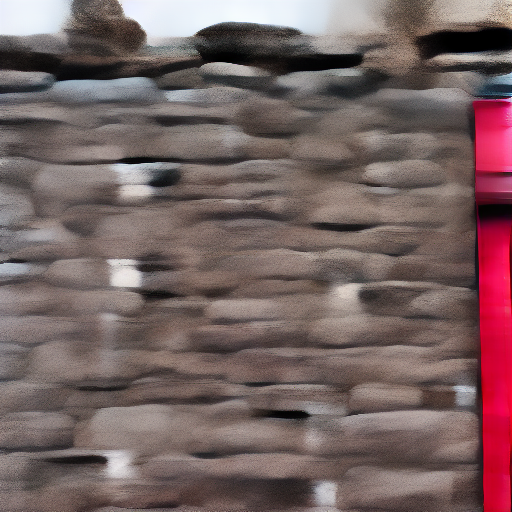} &
    \includegraphics[width=0.10\linewidth]{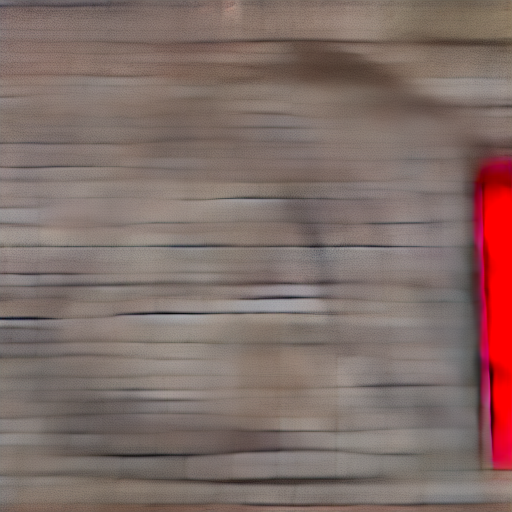} &
    \includegraphics[width=0.10\linewidth]{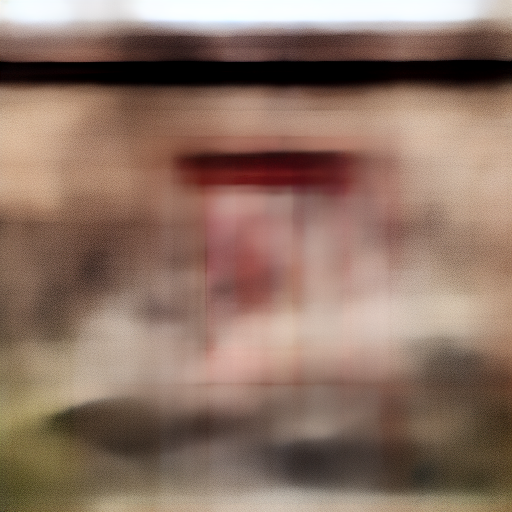} &
    \includegraphics[width=0.10\linewidth]{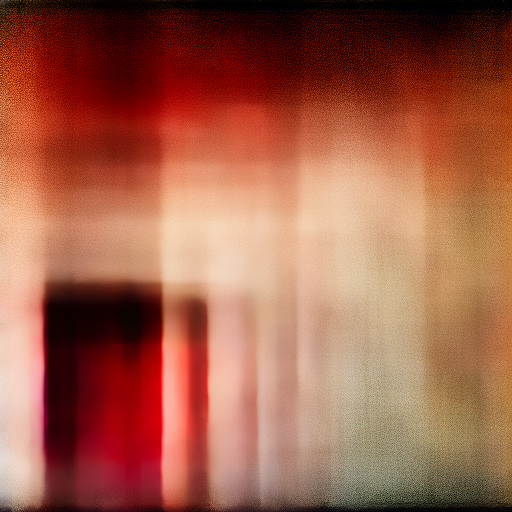} \\[1mm]
    
    \rotatebox{90}{\hspace{0mm} \footnotesize Image-cond.} &
    \includegraphics[width=0.10\linewidth]{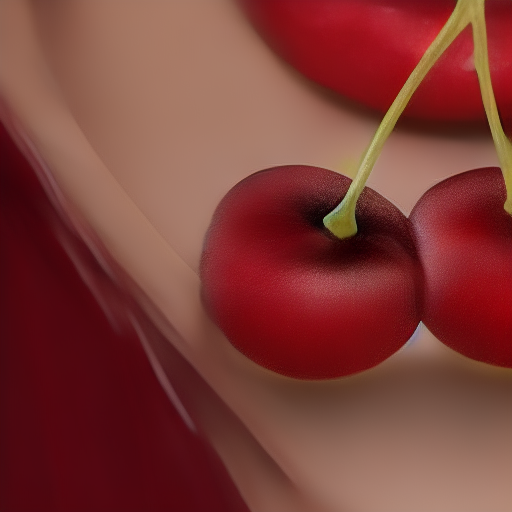} &
    \includegraphics[width=0.10\linewidth]{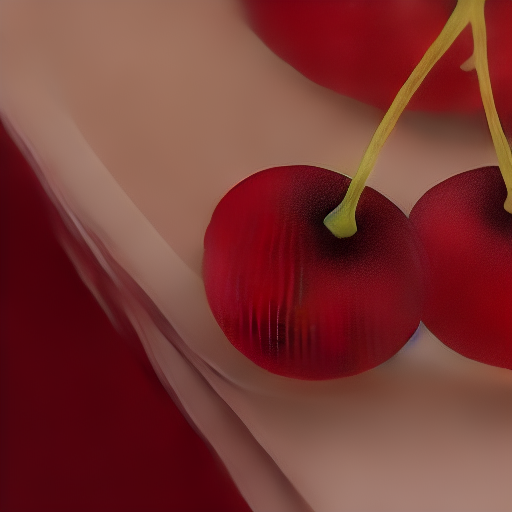} &
    \includegraphics[width=0.10\linewidth]{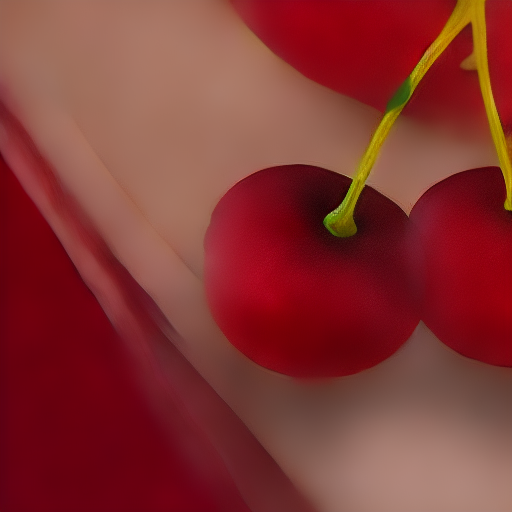} &
    \includegraphics[width=0.10\linewidth]{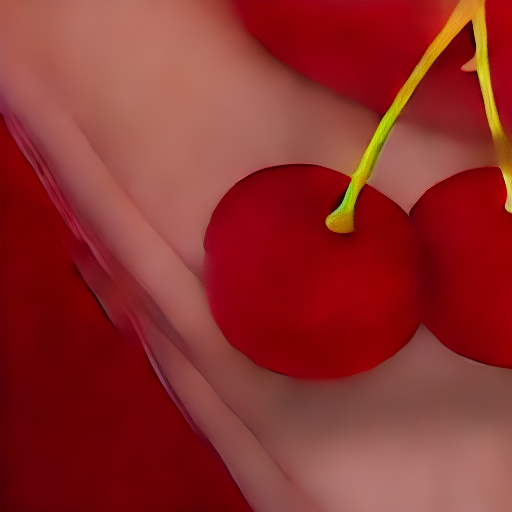} &
    \includegraphics[width=0.10\linewidth]{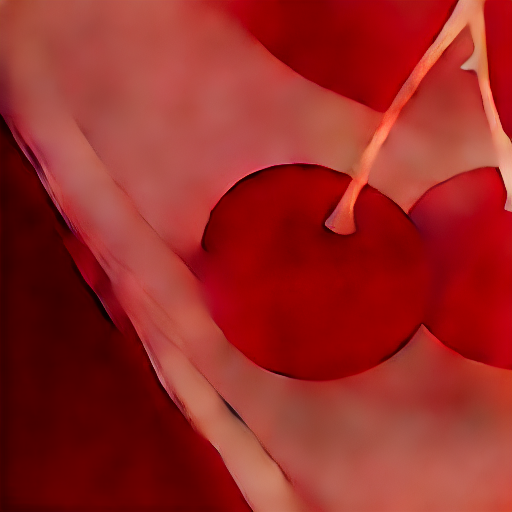} &
    \includegraphics[width=0.10\linewidth]{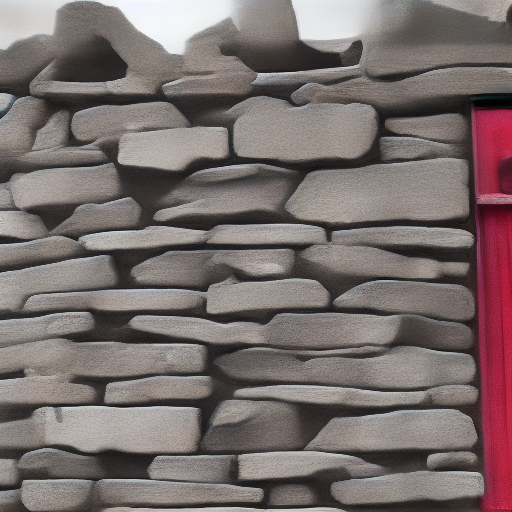} &
    \includegraphics[width=0.10\linewidth]{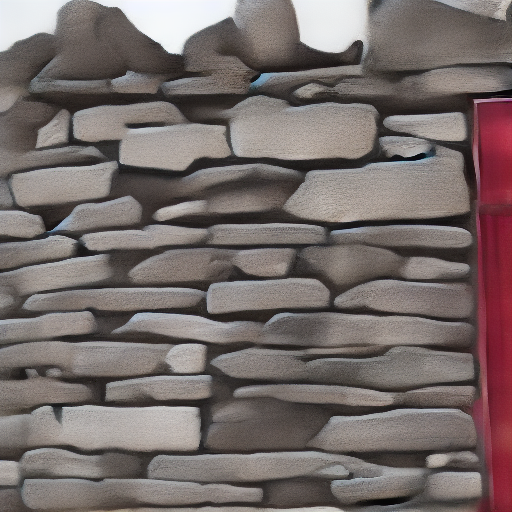} &
    \includegraphics[width=0.10\linewidth]{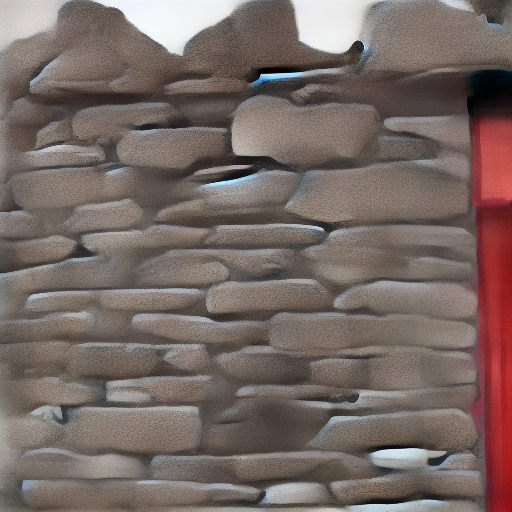} &
    \includegraphics[width=0.10\linewidth]{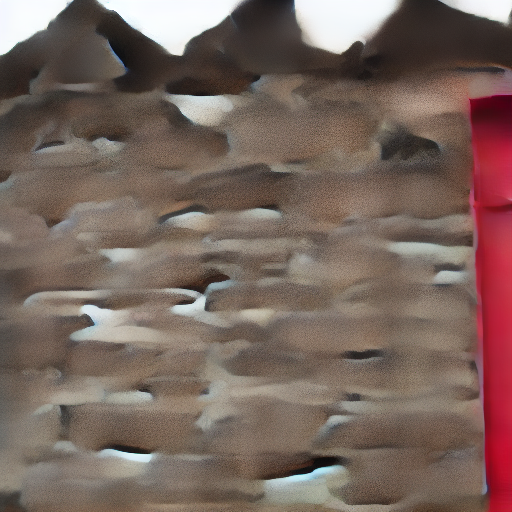} &
    \includegraphics[width=0.10\linewidth]{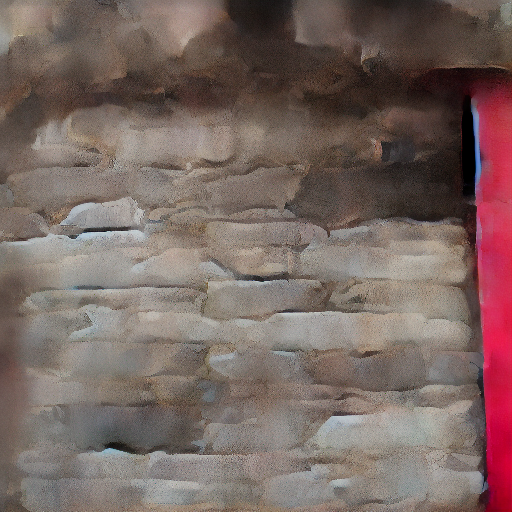} \\

\end{tabular}
}

	\caption{
	Visualization of denoised latents from the teacher diffusion model. 
	We add noise corresponding to timestep $t$ to the ground-truch latent and visualize the model's single-step denoised prediction.
	\textbf{(Text-cond.)} The standard text-conditioned prior struggles to reconstruct the image from noisy latents, especially at large $t$.
	It produces oversaturated colors~(left) and fails to recover edges~(right).
	\textbf{(Image-cond.)} In contrast, our proposed image-conditioned prior, guided by a colormap and Canny edges, provides a much more accurate prediction.}
	It consistently reconstructs latents with faithful color and sharp structural details, demonstrating a more stable and task-aligned generative prior.
	\label{fig:practical}
\end{figure*}

% \textbf{ISR and Real-ISR}
Image Super-Resolution (ISR), which aims to restore a high-quality (HQ) image from its low-quality (LQ) counterpart, is a classical problem in computer vision.
While recent advances in deep learning have significantly improved the ISR performance~\cite{dong2014learning, kim2016accurate, lim2017enhanced, liang2021swinir, guo2024mambair}, they often fail to generalize to the diverse and unknown degradations encountered in real-world scenarios.
To address this limitation, real-world image super-resolution~(Real-ISR)~\cite{zhang2021designing, wang2021real} aims to achieve practical super-resolution by applying significantly more diverse and complex degradation pipelines.
Since training models solely with a pixel-wise reconstruction loss inevitably leads to blurry and oversmoothed results, training frameworks from generative models such as Generative Adversarial Networks~(GANs)~\cite{goodfellow2014generative} and diffusion models~\cite{ho2020denoising,song2021scorebased,rombach2022high} are adopted for Real-ISR.
Both GAN-based methods~\cite{ledig2017photo, wang2018esrgan, wang2021real} and diffusion-based methods~\cite{yue2023resshift, lin2024diffbir} enable the generation of more realistic and sharp images with superior perceptual quality compared to the models trained only with reconstruction loss.

% \textbf{Leveraging generative prior}
The emergence of generative foundation models has opened new avenues for Real-ISR, leveraging the powerful generative priors of pretrained text-to-image diffusion models~\cite{rombach2022high}.
One approach~\cite{wu2024seesr,yu2024scaling} is to adapt pretrained diffusion models to the Real-ISR task by training LoRA~\cite{hu2022lora} or ControlNet~\cite{zhang2023adding}.
This approach preserves the powerful generative priors inherent in the text-to-image models, thereby achieving superior generalization capabilities.
Despite their impressive performance, standard diffusion models for ISR often require high computational cost due to their iterative sampling process.
For efficient inference, researches~\cite{wu2024one, dong2025tsd, li2025one, you2025consistency} have explored distilling generative priors into one-step super-resolution models.
They adopt distillation techniques for diffusion models such as distribution matching~\cite{luo2024diffinstruct,yin2024one} and consistency trajectory matching~\cite{song2023cm,kim2024ctm}.
Among these efficient models, OSEDiff~\cite{wu2024one} regularizes the super-resolution outputs towards the natural image prior embedded within the pretrained diffusion models, facilitated by Variational Score Distillation~(VSD)~\cite{wang2023prolific}. 

% \textbf{Limitation of previous works}
However, many existing one-step Real-ISR methods focus on applying and enhancing distillation techniques developed for general-purpose image generation.
This leads them to overlook a more fundamental aspect: choosing a target manifold that aligns with the characteristics of the Real-ISR task.
Specifically, these methods regularize output towards a manifold conditioned on text prompts.
This approach creates a conceptual mismatch, as the Real-ISR task requires generating an HQ image that is faithful to the LQ input, not just plausible based on a text description.
Moreover, as visualized in \cref{fig:practical}~(Text cond.), the text-conditioned teacher models often reconstruct images with saturated color and blunt boundaries.
This indicates that the generative prior is not only misaligned but also practically flawed for the precise task of image restoration.
An intuitive solution to resolve this mismatch is conditioning the target manifold on the LQ images.
However, we prove that conditioning on the information-dense signal causes VSD~\cite{wang2023prolific} to become numerically unstable and degenerate towards SDS~\cite{poole2022dreamfusion}, thereby harming the distillation performance.

% \textbf{Our contribution}
To address the dilemma between conceptual alignment and distillation stability, we propose Image-Conditioned Manifold~(ICM) regularization.
Our method conditions the target manifold on core structural information, which we compose from a low-resolution colormap and Canny edges.
This combination is specifically designed to resolve the aforementioned practical failure of the text-conditioned prior; the colormap provides global guidance to prevent color shifts, while Canny edges enforce sharp structural details.
We implement this conditioing using a pretrained T2I-Adapter~\cite{mou2023t2iadapter}, and this clearly mines more stable and accurate prior from diffusion models as shown in \cref{fig:practical}~(bottom row).

ICM regularization offers two key advantages.
Conceptually, ICM  provides a regularization manifold that is fundamentally better aligned with the objectives of Real-ISR.
Practically, the structural conditioning improves score estimation accuracy, especially at large diffusion timesteps.
Consequently, this synergy of conceptual alignment and practical stability allows ICM regularization to yield superior one-step diffusion models for Real-ISR.

Our key contributions are summarized as follows:
\begin{itemize}
	\item We propose Image-Conditioned Manifold~(ICM) regularization, highlighting the overlooked importance of the target manifold for regularization. Our key idea is to condition the manifold on image information, which conceptually aligns the generative prior to the Real-ISR task.
	\item We prove that conditioning on information-dense signals causes the VSD loss to degenerate towards SDS, leading to numerical instability. Our method, ICM regularization, resolves this by conditioning only on core structural information. This achieves the desired alignment of the prior while ensuring a stable distillation process.
	\item We show the effectiveness of ICM regularization through extensive experiments on Real-ISR benchmarks, confirming ICM-SR's superior performance, particularly in perceptual quality, validating the practicality of the proposed approach for real-world applications.
\end{itemize}

% !TEX root = ./../main.tex

\section{Related Work}
\label{sec:rel_work}

\subsection{Real-ISR for Perceptual Enhancement}
Early deep learning-based super-resolution methods focused on pixel-wise accuracy, often resulting in blurry outputs for real-world images. 
To address this, the field of Real-World Image Super-Resolution (Real-ISR) emerged, prioritizing perceptual quality~\cite{zhang2021designing, wang2021real}. 
A key advancement was the adoption of Generative Adversarial Networks (GANs), pioneered by SRGAN~\cite{ledig2017photo}. Subsequent methods like Real-ESRGAN~\cite{wang2021real} and BSRGAN~\cite{zhang2021designing} further improved photo-realism by training on more complex and realistic degradation models.

\subsection{Multi-step Diffusion-based Real-ISR}
Recently, diffusion models have set a new standard for perceptual quality in Real-ISR. 
These methods leverage powerful priors from pre-trained text-to-image models, such as Stable Diffusion~\cite{rombach2022high}, by conditioning the generation process on the low-resolution (LQ) input. 
Representative works include StableSR~\cite{wang2024exploiting}, DiffBIR~\cite{lin2024diffbir}, SeeSR~\cite{wu2024seesr}, and SUPIR~\cite{yu2024scaling}, which employ various fine-tuning or adapter-based strategies for conditioning. 
WWhile demonstrating impressive perceptual quality, a major drawback of these multi-step diffusion methods is their high inference cost, requiring numerous sampling steps and resulting in slow processing speeds.

\begin{figure*}[t]
    \centering
    \includegraphics[width=0.98\linewidth]{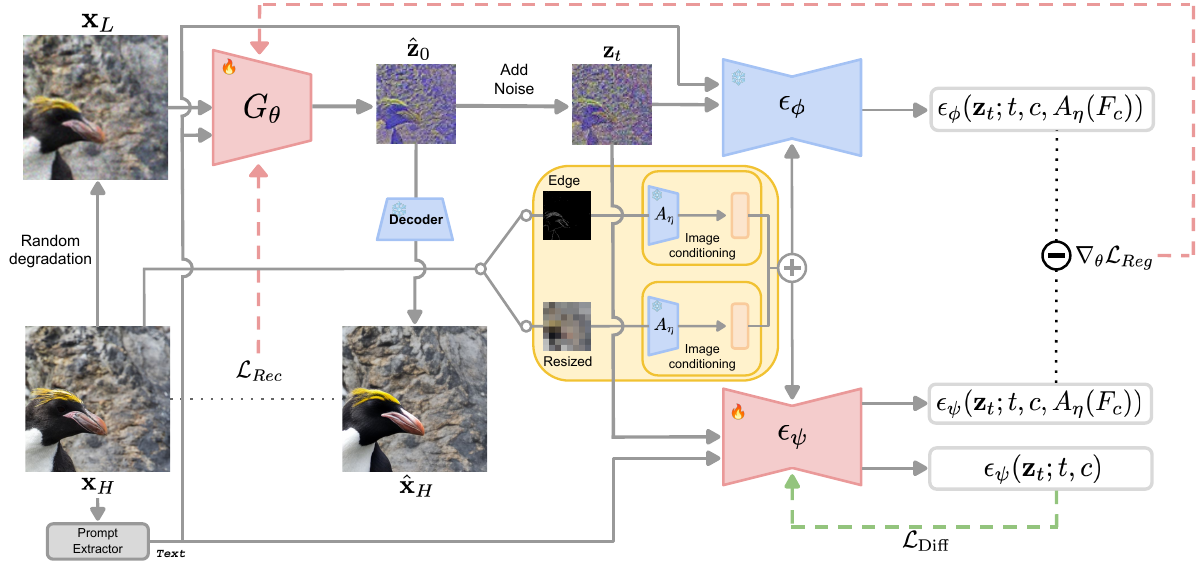}
    \vspace{-2mm}
    \caption{
        Training framework of ICM-SR. 
        Our framework trains a one-step super-resolution generator using two main losses. 
        A reconstruction loss $\mathcal{L}_\text{Rec}$ ensures fidelity to the ground-truth $\mathbf{x}_H$. 
        For realism, a VSD-based regularization loss $\mathcal{L}_\text{Reg}$ is applied, which involves a frozen pre-trained diffusion model $\epsilon_\phi$ and a trainable auxiliary model $\epsilon_\psi$.
        The key innovation of our method is to condition the target manifold on structural information $\F_c$~(\eg, edges, resized image) from the HQ image $\x_H$.
        These conditions are encoded by T2I-Adapter $A_\eta$ and then injected into both $\epsilon_\phi$ and $\epsilon_\psi$ to guide the generator $G_\theta$ towards producing outputs that are not only realistic but also structurally aligned with the target image.}
    \label{fig:method}
\end{figure*}

\subsection{Efficient Diffusion Models for Real-ISR}
The considerable inference time of multi-step diffusion models necessitates acceleration techniques. Distillation has emerged as a primary strategy to reduce sampling steps, often targeting one-step generation for diffusion-based ISR. Noteworthy efficient methods include SinSR~\cite{wang2024sinsr} using consistency preserving distillation, AddSR~\cite{xie2024addsr} employing adversarial diffusion distillation, and OSEDiff~\cite{wu2024one} which proposes a Variational Score Distillation (VSD) loss in the latent space. 
More recently, TSD-SR~\cite{dong2025tsd} introduced Target Score Distillation (TSD) and a Distribution-Aware Sampling Method (DASM) to improve score estimation stability. Although these methods significantly reduce inference steps, challenges persist in achieving perfect score distillation across all timesteps and preventing artifacts, highlighting the need for further improvements in efficient diffusion-based Real-ISR.
% !TEX root = ./../main.tex

\section{Preliminary}
\label{sec:preliminary}
% \label{sec:method}

\paragraph{Real world image super-resolution}
Real-ISR is a task to reconstruct high quality~(HQ) image $\x_H$ from low quality~(LQ) image $\x_L$. 
Given a dataset $\mathcal{D} = \{(\x_L,\x_H)\}_{i=1}^{N}$, a super-resolution model $G_\theta$ is trained to minimize
\begin{align} 
    \mathbb{E}_{(\x_L, \x_H) \sim \mathcal{D}} 
    \left[ 
        \mathcal{L}_\text{Rec}(G_\theta(\x_L), \x_H) +
        \mathcal{L}_\text{Reg}(G_\theta(\x_L))
    \right], \label{eq:isr_loss}
\end{align}
where $\mathcal{L}_\text{Rec}$ is the reconstruction loss such as MSE or LPIPS~\cite{zhang2018unreasonable}, and $\mathcal{L}_\text{Reg}$ is the regularization loss forcing $G_\theta(\x_L)$ to lie on the desired real HQ image manifold.

\paragraph{Regularizing towards a generative prior}
To ensure the output to be photorealistic, the regularization loss $\mathcal{L}_\text{Reg}$ regularizes the model's output distribution towards the powerful generative prior learned by large-scale text-to-image diffusion models.
Following prior distillation works~\cite{wang2023prolific, yin2024one}, OSEDiff~\cite{wu2024one} implements this by defining the desired manifold as the distribution of high-quality images conditioned on text prompts $\cc_t$.
The regularization loss is formalized using the Variational Score Distillation~(VSD) loss:
\begin{align} 
    \mathcal{L}_\text{Reg}&(G_\theta(\x_L)) = \mathcal{L}_\text{VSD}(G_\theta(\x_L), \cc_t) \nonumber \\
	&= \int_0^T w(t) D_\text{KL}(q_t^\theta(\hat{\x}_H|\cc_t)||p_t^\text{real}(\x_H|\cc_t)) dt, \label{eq:isr_reg}
\end{align}
where $\cc_t$ is text prompt describing $\x_H$, $q_t^\theta$ is a perturbed distribution of $\hat{\x}_H=G_\theta(\x_L)$ and $p_t^\text{real}$ is a noisy distribution learned by pretrained diffusion models.
As the computation of $\mathcal{L}_\text{VSD}$ is intractable, the model is optimized by the gradient of $\mathcal{L}_\text{VSD}$ with respect to $\theta$.
The gradient is expressed with two score functions estimated via diffusion models:
\begin{align} 
    & \nabla_\theta \mathcal{L}_\text{VSD}  \nonumber \\
    &= \nabla_\theta \int_0^T \hspace{-2mm} w(t)~ D_\text{KL}(q_t^\theta(\hat{\x}_H|\cc_t)||p_t^\text{real}(\x_H|\cc_t)) dt \nonumber \\
    &= \nabla_\theta \int_0^T \hspace{-2mm} w(t)~ \mathbb{E}_{\hat{\z}_t \sim q_t^\theta}
        \left[\log q_t^\theta(\hat{\z}_t|\cc_t) - \log p_t^\text{real}(\hat{\z}_t|\cc_t)\right] dt \nonumber \\
    &= \int_0^T \hspace{-2.5mm} w(t)~ \mathbb{E}_{\hat{\z}_t} \hspace{-1.5mm} \left[ \left( 
            \nabla \hspace{-0.5mm} \log q_t^\theta(\hat{\z}_t|\cc_t) - \nabla \hspace{-0.5mm} \log p_t^\text{real}(\hat{\z}_t|\cc_t) 
        \right) \frac{\partial \hat{\z}_t}{\partial \theta} \right] 
         dt \nonumber \\
    &\approx \int_0^T \hspace{-2.5mm} w'(t)~ \mathbb{E}_{\hat{\z}_t}
    \left[ \epsilon_\phi(\hat{\z}_t; t, \cc_t) - \epsilon_\psi(\hat{\z}_t; t, \cc_t)\right] \frac{\partial \hat{\z}_t}{\partial \theta} dt,  \label{eq:isr_vsd}
\end{align}
where $\hat{\z}_t$ denotes perturbed latents~(encoded image), $\epsilon_\psi$ is a diffusion model trained to learn distribution of $G_\theta(\x_L)$, and $\epsilon_\phi$ is a pretrained diffusion model that estimates the score function of real-world, high-fidelity image distribution.
% !TEX root = ./../main.tex

\section{Method}
\label{sec:method}

\subsection{Image-conditioned manifold for Real-ISR}
\label{subsec:icm}

% \paragraph{\jun{Ideal} manifold for Real-ISR}
\paragraph{The dilemma of conditioning for Real-ISR}

A core principle for successful regularization is the proper alignment of prior knowledge with the task's objectives.
However, existing regularization methods for Real-ISR have not addressed the choice of the manifold, often defaulting to a text-conditioned manifold inherited from text-to-image models.
This default choice creates a fundamental mismatch with the objective of Real-ISR: generating an HQ image that is a direct enhancement of the given LQ image, not an arbitrary one that merely matches a text description.

This mismatch naturally suggests that a suitable manifold must be conditioned on image information.
While the most straightforward solution is to condition directly on the raw LQ image, our analysis reveals that this approach leads to a critical problem.
Conditioning on such an information-dense signal over-constrains the target manifold, causing the VSD to become numerically unstable and degenerate towards SDS~\cite{poole2022dreamfusion}, thereby harming distillation performance.
We formalize this instability in the following lemma:

 \begin{lemma}
\label{lemma1}
	Let $\cc$ be a strong condition such that the latent variable $\z_0$ is deterministic, \ie, $\z_0|\cc = \mu(\cc)$.
	If the perturbed distribution $q_t(\z_t|\cc)$ is generated by $\z_t = a_t \z_0 + b_t \epsilon$ with $\epsilon \sim \mathcal{N}(\mathbf{0}, \mathbf{I})$, then the auxiliary denoiser $\epsilon_\psi$ collapses to the sampled noise $\epsilon$:
	\begin{align}
		\epsilon_\psi(\z_t; t, \cc) = \epsilon.
	\end{align}
	Consequently, the gradient of the VSD loss $\nabla_\theta \mathcal{L}_\text{VSD}$ degenerates to the gradient of the SDS loss $\nabla_\theta \mathcal{L}_\text{SDS}$.
	In other words, the score prediction difference becomes:
	\begin{align}
		\epsilon_\phi(\z_t; t, \cc) - \epsilon_\psi(\z_t; t, \cc) = \epsilon_\phi(\z_t; t, \cc) - \epsilon,
	\end{align}
	where $\epsilon_\phi$ is the denoiser of real high-quality images.
\end{lemma}
% \begin{proof}
%     Please refer to the supplementary materials.
% \end{proof}
% \renewcommand{\qed}{\unskip\nobreak\quad\qedsymbol}
\begin{proof}
$\z_0|\cc \sim \delta(\z_0 - \mu(\cc))$ implies $\z_t|\cc \sim \mathcal{N} (\mu(\cc), b_t^2 \mathbf{I})$.	
Hence, $\nabla_{\z_t} \! \log q_t (\z_t|\cc) =\! - \frac{\z_t - a_t \mu(\cc)}{b_t^2}$ holds, which leads to
	\begin{align*}
		\epsilon_\psi(\z_t; t, \cc)  
		&= -b_t \nabla_{\z_t} \log q_t (\z_t|\cc)  \\
		&= (-b_t) \times (-\frac{\z_t - a_t \mu(\cc)}{b_t^2}) \\
		&= \frac{a_t \z_0 + b_t \epsilon - a_t \mu(\cc)}{b_t} = \epsilon. 
	\end{align*}
	
	{\vspace{-4mm} \qedhere}
\end{proof}
\noindent An effective manifold for Real-ISR regularization must satisfy two criteria: (1) it must be conditioned on image information to be conceptually aligned, and (2) it must not be overly restrictive to ensure distillation stability.

\vspace{-2mm}
\paragraph{The design of core structural information $\F_c$}

To satisfy both criteria simultaneously, our guiding principle is to extract and condition on the core identity of the target image, while discarding fine-grained, high-density information.
The design of this core structural information, $\F_c$, is motivated by the observed practical failures of the text-conditioned prior.
As visualized in \cref{fig:practical}, the teacher model guided only by text often reconstructs outputs with oversaturated colors and blurred edges.
This indicates that the text-conditioned prior is not only conceptually misaligned but also practically flawed for the image restoration.
To address the failures while adhering to the two criteria, we design the core structural information as a combination of a low-resolution colormap and Canny edges.
The 8x8 colormap guides the teacher model to extract priors aligned with the target's color distribution, while the Canny edges enforce the elicitation of priors rich in structural details.
This combination of sparse yet essential information provides the necessary conceptual alignment from image conditioning~(criterion 1) without the distillation instability caused by the high information density~(criterion 2).

\vspace{-2mm}
\paragraph{Image-conditioned manifold regularization}

We formalize the concept of conditioning on the core structural information with our proposed Image-Conditioned Manifold~(ICM) regularization loss.
This loss regularizes the super-resolution outputs to lie on the manifold conditioned on both the structural information and the text prompt, $p_t^\text{real}(\x_H | \F_c, \cc)$:
\begin{align} \label{eq:icm_reg}
    \hspace{-2mm} \mathcal{L}_\text{ICM} 
    = \hspace{-1mm} \int_0^T \hspace{-2mm} w(t) D_\text{KL}(q_t^\theta(\hat{\x}_H|\F_c, \cc_t)||p_t^\text{real}(\x_H|\F_c, \cc_t)) dt.
\end{align}
The gradient of this loss with respect to $\theta$ can be derived similarly to \cref{eq:isr_vsd}, yielding:
\begin{align} 
    \nabla_\theta & \mathcal{L}_\text{ICM} \label{eq:icm_loss} \\
    &\approx \int_0^T \hspace{-2mm} w'(t)~ \mathbb{E}_{\hat{\z}_t \sim q_t^\theta} 
    \left[ \epsilon_\phi(\hat{\z}_t; t, \cc_t, \F_c) - \epsilon_\psi(\hat{\z}_t; t, \cc_t, \F_c) \right] dt.  \nonumber  
\end{align}

\begin{algorithm}[t]
\caption{ICM-SR}\label{alg:icm-sr}
    % !TEX root = ./../main.tex

\begin{algorithmic}[1]
	\STATE \textbf{Require}: $G_\theta$, $\epsilon_\phi$, $\epsilon_\psi$, $A_\eta$, Dec, Scheduler, $F$
	\WHILE{train}
		\STATE $(\x_L, \x_H, \cc_t) \sim \mathcal{D}$
		\STATE $\hat{\z}_0 \leftarrow G_\theta(\x_L)$, $\F_c \leftarrow F(\x_H)$
		
		\vspace{2mm}
		\STATE \texttt{/* Compute reconstruction loss}
		\STATE $\hat{\x}_H \leftarrow \text{Dec}(\hat{\z}_0)$
		\STATE $\mathcal{L}_\text{Rec} \leftarrow \mathcal{L}_\text{2}(\hat{\x}_H, \x_H) + \mathcal{L}_\text{LPIPS}(\hat{\x}_H, \x_H)$
		
		\vspace{2mm}
		\STATE \texttt{/* Compute regularization gradient}
		\STATE $t \sim \mathcal{U}(20,980), \epsilon \sim \mathcal{N}(\mathrm{0}, \mathrm{I})$
		\STATE $a_t, b_t \leftarrow$ Scheduler(t)
		\STATE $\hat{\z}_t \leftarrow a_t \hat{\z}_0 + b_t \epsilon$
		\STATE $\epsilon_\text{fake} \leftarrow \text{stopgrad}(\epsilon_\psi(\hat{\z}_t;t,\cc_t,A_\eta(\F_c))$ 
		\STATE $\epsilon_\text{real} \leftarrow \text{stopgrad}(\text{cfg}(\epsilon_\phi(\hat{\z}_t;t,\cc_t,A_\eta(\F_c)))$
		\STATE $\nabla_\theta\mathcal{L}_\text{Reg} \leftarrow w(t) (\epsilon_\text{fake} - \epsilon_\text{real})$
		\vspace{2mm}
		\STATE \texttt{/* Compuate auxilary diffusion loss}%
		\STATE $\hat{\z}_t \leftarrow \text{stopgrad}(\hat{\z}_t)$
		\STATE $\mathcal{L}_\text{Aux} \leftarrow ||\epsilon_\psi(\hat{\z}_t;t,\cc_t) -\epsilon||_2^2$ 
		\vspace{2mm}
		\STATE \texttt{/* Update generator $G_\theta$ and auxiliary diffusion model $\epsilon_\psi$}
%		\STATE \texttt{/* Update $G_\theta$ and $\epsilon_\psi$}
		\STATE Update $\theta$ with $\mathcal{L}_\text{Rec}$ and $\nabla_\theta \mathcal{L}_\text{Reg}$
		\STATE Update $\psi$ with $\mathcal{L}_\text{Aux}$
	\ENDWHILE 
\end{algorithmic}

\end{algorithm}

\subsection{Training framework of ICM-SR}

The overall training framework of ICM-SR is designed to regularize the super-resolution output towards the image-conditioned manifold.
As depicted in \cref{fig:method} and \cref{alg:icm-sr}, the process involves training $G_\theta$ with two main losses and training an auxiliary diffusion model, $\epsilon_\psi$, which learns the distribution of current super-resolution output, $G_\theta(\x_L)$.
The key components of our framework are detailed below.

\subsubsection{Training super-resolution model $G_\theta$}

The super-resolution model $G_\theta$ is trained with a combined loss function, which consists of a reconstruction loss and a regularization loss.

\vspace{-2mm}
\paragraph{Reconstruction loss}
Following OSEDiff~\cite{wu2024one}, the reconstruction loss measures the pixel-wise and perceptual difference between the generated and ground-truth images. 
It is defined as the sum of $\ell_2$ and LPIPS losses: 
\begin{align}
 	\mathcal{L}_\text{Rec}(\x_H, \hat{\x}_H) = ||\hat{\x}_H - \x_H||_2^2 + \mathcal{L}_\text{LPIPS}(\hat{\x}_H, \x_H), \label{eq:loss_recon}
\end{align}
for $\hat{x}_H = \text{Dec}(G_\theta(\x_L))$.

\vspace{-2mm}
\paragraph{ICM regularization loss}
The ICM regularization encourages the output to lie on the proposed manifold, $\x_H|\F_c,\cc_t$.
As derived in \cref{eq:icm_loss}, the gradient used for the optimization is: 
\begin{align}
	\nabla_\theta &\mathcal{L}_\text{Reg} (\hat{\z}_t,\x_L,t,\cc_t, \F_c)  \label{eq:loss_reg} \\ 
	&= w(t) \left[ \epsilon_\psi(\hat{\z}_t; t, \cc_t, A_\eta(\F_c)) - \epsilon_\phi(\hat{\z}_t; t, \cc_t, A_\eta(\F_c)) \right], \nonumber 
\end{align}
where $\hat{\z}_t = a_t \hat{\z}_0 + b_t \epsilon$ for $a_t$ and $b_t$ are predefined noise scheduling, $\epsilon_\phi$ is the pretrained teacher model, $\epsilon_\psi$ is the auxiliary student model, and the core structural information $\F_c$ is injected using the T2I-Adapter $A_\eta$.

\begin{table*}[ht]
	\centering
	\caption{Quantitative comparison with state-of-the-art one-step super-resolution methods on both synthetic~(DIV2K-Val) and real-world benchmarks~(DrealSR, RealSR). 
	`$\dagger$' indicates models trained by us for fair comparison.
	The best and second best results are highlighted in \textcolor{red}{\textbf{red}} and \textcolor{blue}{blue}, respectively.}
	% !TEX root = ./../main.tex

\renewcommand{\arraystretch}{1} % Increase row spacing
\setlength\tabcolsep{5pt}
\scalebox{0.85}{
  \begin{tabular}{cl|ccc|cccccc|cc}
  \toprule
    \multirow{2}{*}{Datasets} & \multirow{2}{*}{Methods} & \multicolumn{3}{c|}{Perceptual w/ ref.} & \multicolumn{6}{c|}{ Perceptual w/o ref.} & \multicolumn{2}{c}{Fidelity w/ ref.} \\
    {} & {}           & {LPIPS$\downarrow$}      & {DISTS$\downarrow$}      & {FID$\downarrow$}        & {NIQE$\downarrow$}       & {MUSIQ$\uparrow$}       & {MANIQ$\uparrow$}        & {CLIPIQ$\uparrow$}       & {TOPIQ$\uparrow$}        & {LIQE$\uparrow$}         & {PSNR$\uparrow$}        & {SSIM$\uparrow$}         \\ \midrule
    
    % {Datasets} & {Methods}           & {LPIPS$\downarrow$}      & {DISTS$\downarrow$}      & {FID$\downarrow$}        & {NIQE$\downarrow$}       & {MUSIQ$\uparrow$}       & {MANIQ$\uparrow$}        & {CLIPIQ$\uparrow$}       & {TOPIQ$\uparrow$}        & {LIQE$\uparrow$}         & {PSNR$\uparrow$}        & {SSIM$\uparrow$}         \\ \hline
    \multirow{5}{*}{DIV2K-Val}
               & SinSR               & 0.3220                   & 0.2044                   & 37.29                    & 5.7861                   & 63.28                   & 0.5411                   & \textcolor{red}{\textbf{0.6537}}  & 0.5796                   & 3.5015                   & \textcolor{blue}{24.29} & 0.6012                   \\
               & AddSR               & 0.3779                   & 0.2174                   & 30.58                    & 5.2324                   & 58.91                   & 0.5260                   & 0.5337                   & 0.5218                   & 3.3877                   & 23.10                   & 0.5928                   \\
               & OSEDiff${}^\dagger$ & 0.2847                   & 0.1905                   & 26.15                    & \textcolor{blue}{4.4918} & \textcolor{blue}{67.73} & \textcolor{blue}{0.6081} & 0.6394                   & 0.6014                   & \textcolor{blue}{4.1704} & 23.40                   & 0.6160                   \\
               & TSD-SR${}^\dagger$  & \textcolor{red}{\textbf{0.2759}}  & \textcolor{blue}{0.1894} & \textcolor{blue}{25.45}  & 4.6859                   & 65.06                   & 0.5935                   & 0.6306                   & \textcolor{red}{\textbf{0.6252}}  & 3.6920                   & \textcolor{red}{\textbf{24.68}}  & \textcolor{red}{\textbf{0.6257}}  \\
               & ICM-SR              & \textcolor{blue}{0.2799} & \textcolor{red}{\textbf{0.1861}}  & \textcolor{red}{\textbf{24.72}}   & \textcolor{red}{\textbf{4.4411}}  & \textcolor{red}{\textbf{68.00}}  & \textcolor{red}{\textbf{0.6169}}  & \textcolor{blue}{0.6440} & \textcolor{blue}{0.6138} & \textcolor{red}{\textbf{4.2094}}  & 23.77                   & \textcolor{blue}{0.6173} \\ \midrule

    \multirow{5}{*}{DrealSR}
               & SinSR               & 0.3537                   & 0.2495                   & 177.37                   & 6.7022                   & 57.14                   & 0.5021                   & 0.6590                   & 0.5369                   & 3.2267                   & \textcolor{blue}{28.12} & 0.7533                   \\
               & AddSR               & 0.3079                   & 0.2207                   & 154.24                   & 7.4512                   & 53.47                   & 0.4770                   & 0.5451                   & 0.4840                   & 2.8429                   & 27.92                   & \textcolor{blue}{0.7880} \\
               & OSEDiff${}^\dagger$ & 0.2974                   & 0.2183                   & 130.50                   & \textcolor{blue}{6.4852} & \textcolor{blue}{65.46} & \textcolor{blue}{0.5975} & \textcolor{blue}{0.6840} & 0.5961                   & \textcolor{blue}{4.0183} & 25.57                   & 0.7705                   \\
               & TSD-SR${}^\dagger$  & \textcolor{red}{\textbf{0.2869}}  & \textcolor{red}{\textbf{0.2134}}  & \textcolor{blue}{127.29} & 6.5985                   & 62.13                   & 0.5741                   & 0.6614                   & \textcolor{blue}{0.6028} & 3.5815                   & \textcolor{red}{\textbf{28.25}}  & \textcolor{red}{\textbf{0.7885}}  \\
               & ICM-SR              & \textcolor{blue}{0.2871} & \textcolor{blue}{0.2142} & \textcolor{red}{\textbf{125.30}}  & \textcolor{red}{\textbf{6.4163}}  & \textcolor{red}{\textbf{65.96}}  & \textcolor{red}{\textbf{0.6051}}  & \textcolor{red}{\textbf{0.6929}}  & \textcolor{red}{\textbf{0.6088}}  & \textcolor{red}{\textbf{4.0977}}  & 26.85                   & 0.7763                   \\ \midrule

    \multirow{5}{*}{RealSR}
               & SinSR               & 0.3050                   & 0.2325                   & 135.51                   & 6.0516                   & 60.82                   & 0.5423                   & 0.6212                   & 0.5321                   & 3.1935                   & \textcolor{red}{\textbf{26.15}}  & \textcolor{red}{\textbf{0.7385}}  \\
               & AddSR               & 0.3141                   & 0.2198                   & 135.58                 & {6.1957}                 & 62.65                   & 0.5620                   & 0.5141                   & 0.5532                   & 3.3821                   & 24.22                   & 0.7032                   \\
               & OSEDiff${}^\dagger$ & \textcolor{blue}{0.2637} & \textcolor{blue}{0.2023} & \textcolor{blue}{112.26} & \textcolor{blue}{5.5930} & \textcolor{blue}{68.16} & \textcolor{blue}{0.6246} & \textcolor{blue}{0.6271} & 0.5998                   & \textcolor{blue}{4.0785} & 24.22                   & 0.7276                   \\
               & TSD-SR${}^\dagger$  & 0.2699                   & 0.2049                   & 116.17                   & 5.6907                   & 66.05                   & 0.6057                   & 0.6106                   & \textcolor{red}{\textbf{0.6210}}  & 3.6515                   & \textcolor{blue}{25.92} & \textcolor{blue}{0.7366} \\
               & ICM-SR              & \textcolor{red}{\textbf{0.2611}}  & \textcolor{red}{\textbf{0.2009}}  & \textcolor{red}{\textbf{108.74}}  & \textcolor{red}{\textbf{5.5842}}  & \textcolor{red}{\textbf{68.59}}  & \textcolor{red}{\textbf{0.6288}}  & \textcolor{red}{\textbf{0.6360}}  & \textcolor{blue}{0.6094} & \textcolor{red}{\textbf{4.1336}}  & 24.99                   & 0.7309                   \\ \bottomrule
  \end{tabular}%
}

	\label{tab:main_comp}
	% \vspace{-2mm}
  \end{table*}
\subsubsection{Training auxiliary diffusion model $\epsilon_\psi$}

The auxiliary diffusion model with trainable LoRA~\cite{hu2022lora} learns the distribution of the generator's current output.
Therefore, its objective is solely based on this output without conditioning on the external map:
\begin{align}
	\mathcal{L}_\text{Aux} &(\hat{\z}_t,t,\cc_t)
= ||\epsilon_\psi(\hat{\z}_t;t,\cc_t) - \epsilon||_2^2. \label{eq:loss_aux}
\end{align}

% !TEX root = ./../main.tex

\section{Experiment}
\label{sec:exp}

\subsection{Experimental Settings}

\paragraph*{Training details}
\label{subsec:exp_detials}
For training dataset, we utilize a combined dataset comprising DIV2K~\cite{agustsson2017ntire} and LSDIR~\cite{li2023lsdir}.
Following common practice in real-world super-resolution, low-quality~(LQ) images are synthesized from their high-quality~(HQ) counterparts using the degradation pipeline of Real-ESRGAN~\cite{wang2021real}.
During the training process, we randomly crop $512 \times 512$ patches from the HR images.
The model is optimized using the AdamW optimizer~\cite{loshchilov2018decoupled} for a total of 100K iterations with a batch size of 4. The initial learning rate is set to $5 \times 10^{-5}$ and is adjusted using a cosine learning rate scheduler. 
For the LoRA~\cite{hu2022lora} components integrated into our model, we set the rank to 4.
We utilize pretrained Stable Diffusion v1-4~\footnote{\url{https://huggingface.co/CompVis/stable-diffusion-v1-4}} and T2I-Adapter~\footnote{\url{https://huggingface.co/TencentARC/t2iadapter_color_sd14v1}} trained by TecentARC for ICM-SR.
The conditioning function, $\F_c$, generates guidance either through Canny edge detection or by downsampling an HQ image.
For the downsampling condition, the image is first resized to $8\times 8$ pixels and subsequently upsampled to $512 \times 512$ pixels via nearest-neighbor interpolation to align with the T2I-Adapter's input requirements. 
All trainings were conducted using four NVIDIA RTX A6000 GPUs.

\begin{figure*}[t]
	\centering
	% !TEX root = ./../main.tex

  \scalebox{0.985}{
    \begin{tabular}{@{\hspace{-0.5em}}c@{\hspace{0.3em}}c@{\hspace{0.3em}}c@{\hspace{0.3em}}c@{\hspace{0.3em}}c@{}}

      \parbox[t]{0.18\textwidth}{
        \centering
        \includegraphics[width=\linewidth]{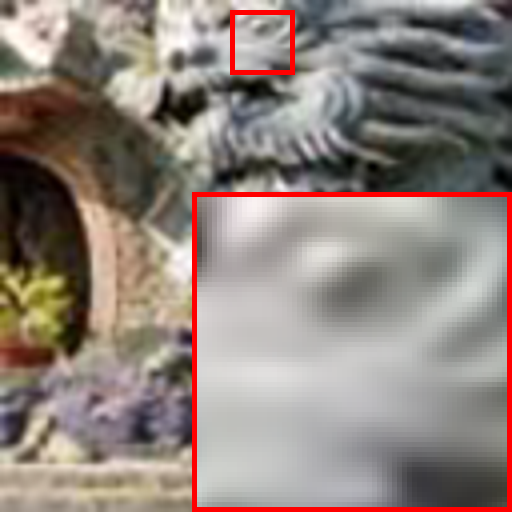}
        \vspace{-7mm}
        \caption*{ \fontsize{8.7pt}{8.7pt}\selectfont LQ}
        \vspace{0mm}
      }  &
      \parbox[t]{0.18\textwidth}{
        \centering
        \includegraphics[width=\linewidth]{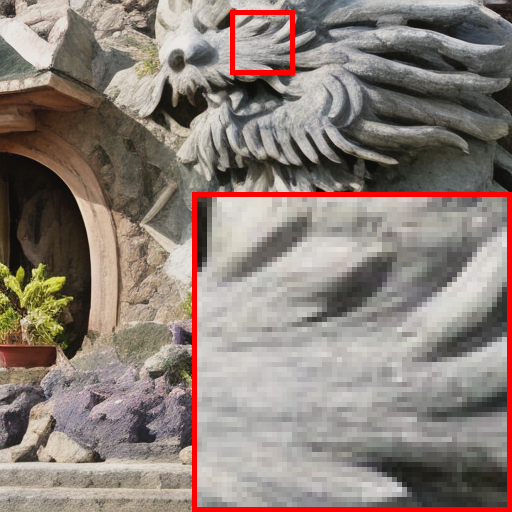}
        \vspace{-7mm}
        \caption*{ \fontsize{8.7pt}{8.7pt}\selectfont OSEDiff}
      }  &
      \parbox[t]{0.18\textwidth}{
        \centering
        \includegraphics[width=\linewidth]{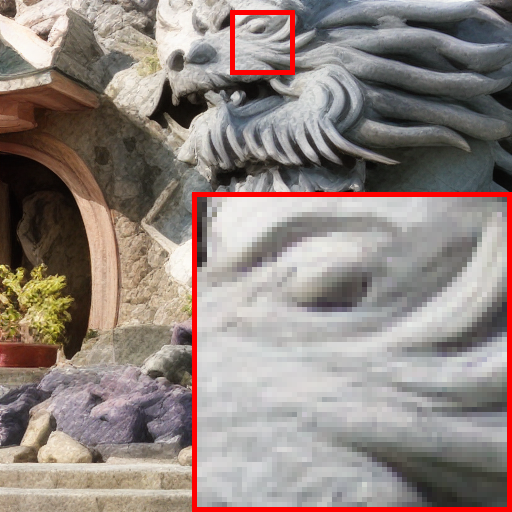}
        \vspace{-7mm}
        \caption*{ \fontsize{8.7pt}{8.7pt}\selectfont TSD-SR}
      }  &
      \parbox[t]{0.18\textwidth}{
        \centering
        \includegraphics[width=\linewidth]{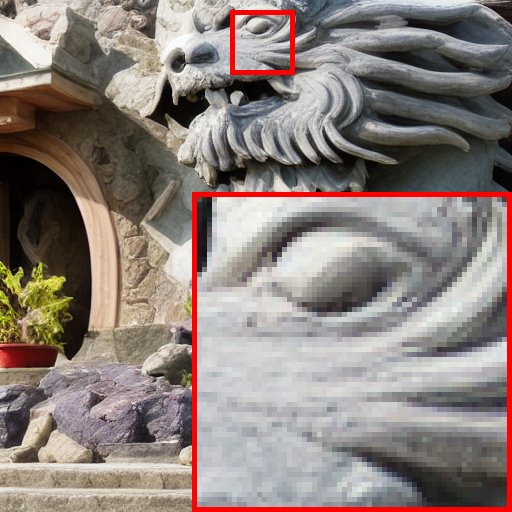}
        \vspace{-7mm}
        \caption*{ \fontsize{8.7pt}{8.7pt}\selectfont ICM-SR (Ours)}
      }  &
      \parbox[t]{0.18\textwidth}{
        \centering
        \includegraphics[width=\linewidth]{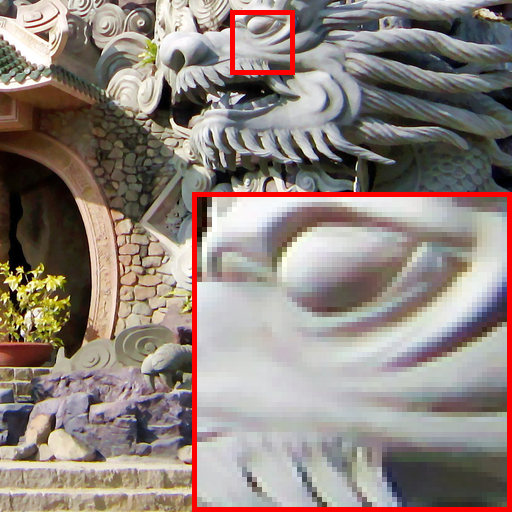}
        \vspace{-7mm}
        \caption*{ \fontsize{8.7pt}{8.7pt}\selectfont GT}
      } 
      \vspace{-2mm}
    \end{tabular}
  }
  \vspace{2mm}

  \scalebox{0.985}{
    \begin{tabular}{@{\hspace{-0.5em}}c@{\hspace{0.3em}}c@{\hspace{0.3em}}c@{\hspace{0.3em}}c@{\hspace{0.3em}}c@{}}
      \parbox[t]{0.18\textwidth}{
        \centering
        \includegraphics[width=\linewidth]{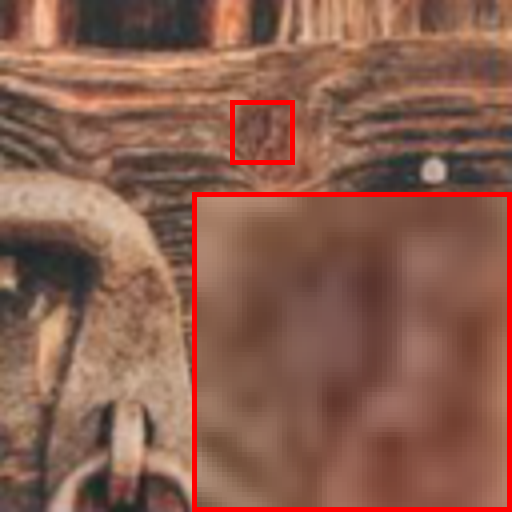}
        \vspace{-7mm}
        \caption*{ \fontsize{8.7pt}{8.7pt}\selectfont LQ}
        \vspace{0mm}
      }  &
      \parbox[t]{0.18\textwidth}{
        \centering
        \includegraphics[width=\linewidth]{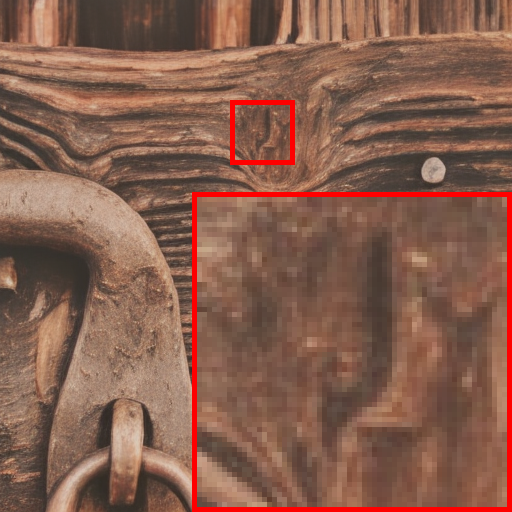}
        \vspace{-7mm}
        \caption*{ \fontsize{8.7pt}{8.7pt}\selectfont OSEDiff}
      }  &
      \parbox[t]{0.18\textwidth}{
        \centering
        \includegraphics[width=\linewidth]{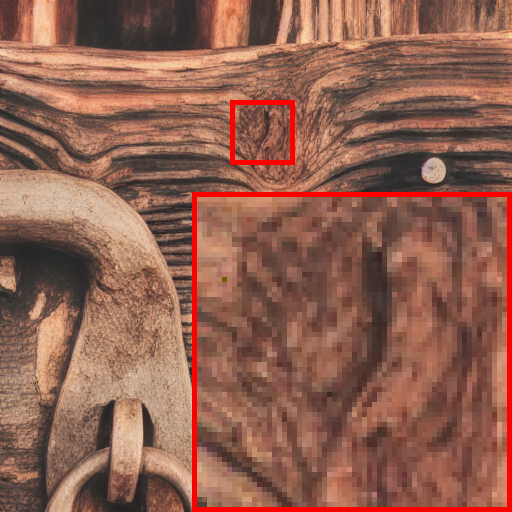}
        \vspace{-7mm}
        \caption*{ \fontsize{8.7pt}{8.7pt}\selectfont TSD-SR}
      }  &
      \parbox[t]{0.18\textwidth}{
        \centering
        \includegraphics[width=\linewidth]{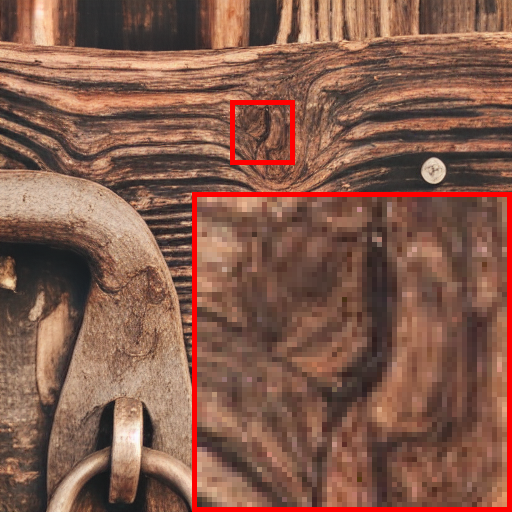}
        \vspace{-7mm}
        \caption*{ \fontsize{8.7pt}{8.7pt}\selectfont ICM-SR (Ours)}
      }  &
      \parbox[t]{0.18\textwidth}{
        \centering
        \includegraphics[width=\linewidth]{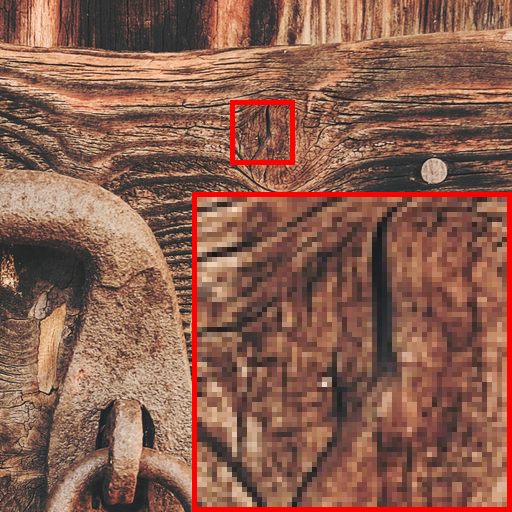}
        \vspace{-7mm}
        \caption*{ \fontsize{8.7pt}{8.7pt}\selectfont GT}
      } 
      \vspace{-2mm}
    \end{tabular}
  }
  % \scalebox{0.89}{
  %   \begin{tabular}{@{\hspace{-0.5em}}c@{\hspace{0.3em}}c@{\hspace{0.3em}}c@{\hspace{0.3em}}c@{\hspace{0.3em}}c@{}}
  %     \parbox[t]{0.18\textwidth}{
  %       \centering
  %       \includegraphics[width=\linewidth]{figures/aaai_qual/div2k/0885_pch_00013/LQ.png}
  %       \vspace{-7mm}
  %       \caption*{ \fontsize{8.7pt}{8.7pt}\selectfont LQ}
  %       \vspace{0mm}
  %     }  &
  %     \parbox[t]{0.18\textwidth}{
  %       \centering
  %       \includegraphics[width=\linewidth]{figures/aaai_qual/div2k/0885_pch_00013/OSEDiff.png}
  %       \vspace{-7mm}
  %       \caption*{ \fontsize{8.7pt}{8.7pt}\selectfont OSEDiff}
  %     }  &
  %     \parbox[t]{0.18\textwidth}{
  %       \centering
  %       \includegraphics[width=\linewidth]{figures/aaai_qual/div2k/0885_pch_00013/TSD-SR.png}
  %       \vspace{-7mm}
  %       \caption*{ \fontsize{8.7pt}{8.7pt}\selectfont TSD-SR}
  %     }  &
  %     \parbox[t]{0.18\textwidth}{
  %       \centering
  %       \includegraphics[width=\linewidth]{figures/aaai_qual/div2k/0885_pch_00013/Ours-multi.png}
  %       \vspace{-7mm}
  %       \caption*{ \fontsize{8.7pt}{8.7pt}\selectfont ICM-SR (Ours)}
  %     }  &
  %     \parbox[t]{0.18\textwidth}{
  %       \centering
  %       \includegraphics[width=\linewidth]{figures/aaai_qual/div2k/0885_pch_00013/GT.png}
  %       \vspace{-7mm}
  %       \caption*{ \fontsize{8.7pt}{8.7pt}\selectfont GT}
  %     } 
  %     % \vspace{-2mm}
  %   \end{tabular}
  % }

  \caption{Qualitative results of our method compared to OSEDiff and TSD-SR on the Div2k validation dataset. Our method demonstrates superior performance in recovering fine details. Zoom in for better visualization.}
	\label{fig:qual_osediff}
\end{figure*}

\begin{figure*}[t]
  \centering
  \scalebox{0.92}{
  \begin{tabular}{@{}c@{\hspace{0.3em}}c@{\hspace{0.3em}}c@{\hspace{0.3em}}c@{\hspace{0.3em}}c@{\hspace{0.3em}}c@{\hspace{0.3em}}c@{}}
    \multirow{2}{*}[5.93em]{\parbox[t]{0.29465\textwidth}{
        \centering
        \vspace{-2.68mm}
        \includegraphics[width=\linewidth]{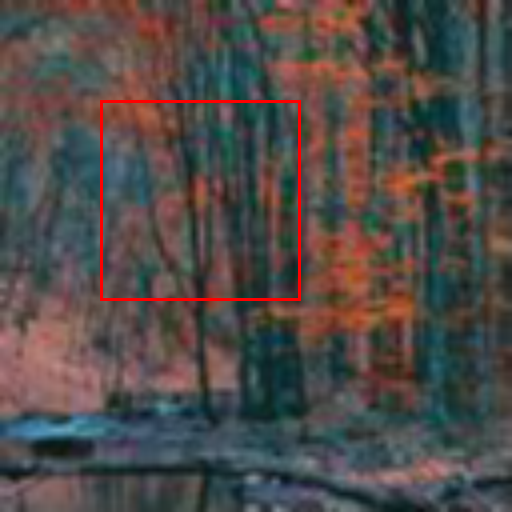}
        \vspace{-7mm}
        \caption*{ \fontsize{8.7pt}{8.7pt}\selectfont LQ}
    }} &
    \parbox[t]{0.135\textwidth}{
      \centering
      \includegraphics[width=\linewidth]{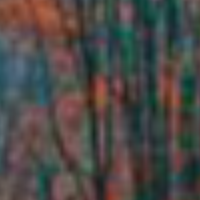}
      \vspace{-7mm}
      \caption*{ \fontsize{8.7pt}{8.7pt}\selectfont Zoomed LQ}
      \vspace{0mm}
    }  &
    \parbox[t]{0.135\textwidth}{
      \centering
      \includegraphics[width=\linewidth]{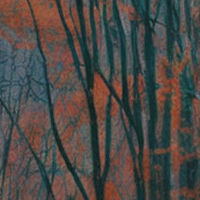}
      \vspace{-7mm}
      \caption*{ \fontsize{8.7pt}{8.7pt}\selectfont StableSR-s200}
    }  &
    \parbox[t]{0.135\textwidth}{
      \centering
      \includegraphics[width=\linewidth]{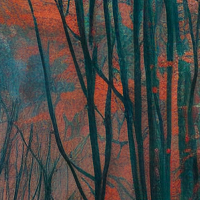}
      \vspace{-7mm}
      \caption*{ \fontsize{8.7pt}{8.7pt}\selectfont DiffBIR-s50}
    }  &
    \parbox[t]{0.135\textwidth}{
      \centering
      \includegraphics[width=\linewidth]{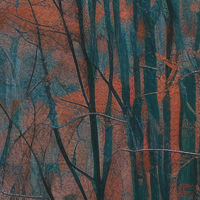}
      \vspace{-7mm}
      \caption*{ \fontsize{8.7pt}{8.7pt}\selectfont SeeSR-s50}
    }  & \hspace{-0.5em}
    \parbox[t]{0.135\textwidth}{
      \centering
      \includegraphics[width=\linewidth]{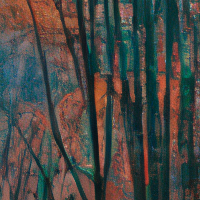}
      \vspace{-7mm}
      \caption*{ \fontsize{8.7pt}{8.7pt}\selectfont Resshift-s15}
    }

    \\

    \vspace{2mm}
       &
    \parbox[t]{0.135\textwidth}{
      \centering
      \includegraphics[width=\linewidth]{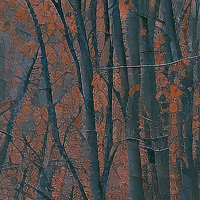}
      \vspace{-7mm}
      \caption*{ \fontsize{8.7pt}{8.7pt}\selectfont AddSR-s4}
    }  &
    \parbox[t]{0.135\textwidth}{
      \centering
      \includegraphics[width=\linewidth]{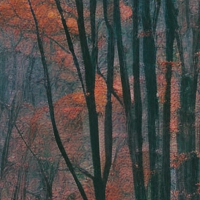}
      \vspace{-7mm}
      \caption*{ \fontsize{8.7pt}{8.7pt}\selectfont OSEDiff-s1}
    }  &
    \parbox[t]{0.135\textwidth}{
      \centering
      \includegraphics[width=\linewidth]{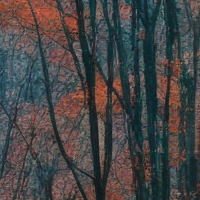}
      \vspace{-7mm}
      \caption*{ \fontsize{8.7pt}{8.7pt}\selectfont TSD-SR-s1}
    }  &
    \parbox[t]{0.135\textwidth}{
      \centering
      \includegraphics[width=\linewidth]{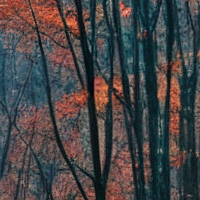}
      \vspace{-7mm}
      \caption*{ \fontsize{8.7pt}{8.7pt}\selectfont ICM-SR-s1}
    }  &
    \parbox[t]{0.135\textwidth}{
      \centering
      \includegraphics[width=\linewidth]{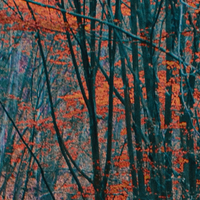}
      \vspace{-7mm}
      \caption*{ \fontsize{8.7pt}{8.7pt}\selectfont GT}
    }
    % \vspace{-1.8mm}
  \end{tabular}
}
  \scalebox{0.92}{
  \begin{tabular}{@{}c@{\hspace{0.3em}}c@{\hspace{0.3em}}c@{\hspace{0.3em}}c@{\hspace{0.3em}}c@{\hspace{0.3em}}c@{\hspace{0.3em}}c@{}}
    \multirow{2}{*}[5.93em]{\parbox[t]{0.29465\textwidth}{
        \centering
        \vspace{-2.68mm}
        \includegraphics[width=\linewidth]{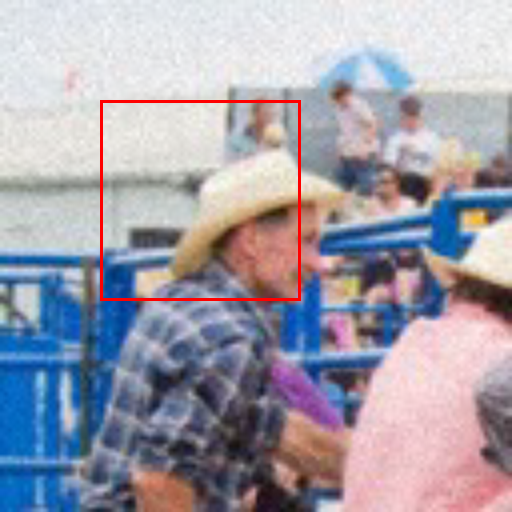}
        \vspace{-7mm}
        \caption*{ \fontsize{8.7pt}{8.7pt}\selectfont LQ}
    }} &
    \parbox[t]{0.135\textwidth}{
      \centering
      \includegraphics[width=\linewidth]{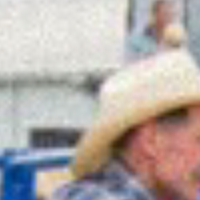}
      \vspace{-7mm}
      \caption*{ \fontsize{8.7pt}{8.7pt}\selectfont Zoomed LQ}
      \vspace{0mm}
    }  &
    \parbox[t]{0.135\textwidth}{
      \centering
      \includegraphics[width=\linewidth]{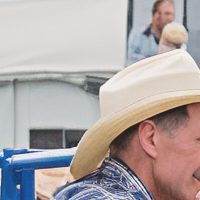}
      \vspace{-7mm}
      \caption*{ \fontsize{8.7pt}{8.7pt}\selectfont StableSR-s200}
    }  &
    \parbox[t]{0.135\textwidth}{
      \centering
      \includegraphics[width=\linewidth]{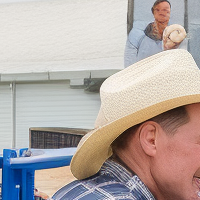}
      \vspace{-7mm}
      \caption*{ \fontsize{8.7pt}{8.7pt}\selectfont DiffBIR-s50}
    }  &
    \parbox[t]{0.135\textwidth}{
      \centering
      \includegraphics[width=\linewidth]{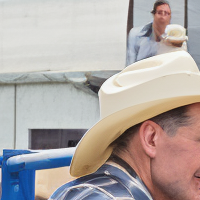}
      \vspace{-7mm}
      \caption*{ \fontsize{8.7pt}{8.7pt}\selectfont SeeSR-s50}
    }  & \hspace{-0.5em}
    \parbox[t]{0.135\textwidth}{
      \centering
      \includegraphics[width=\linewidth]{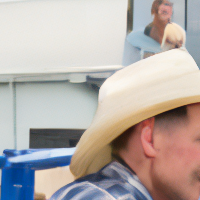}
      \vspace{-7mm}
      \caption*{ \fontsize{8.7pt}{8.7pt}\selectfont Resshift-s15}
    }

    \\

    \vspace{2mm}
       &
    \parbox[t]{0.135\textwidth}{
      \centering
      \includegraphics[width=\linewidth]{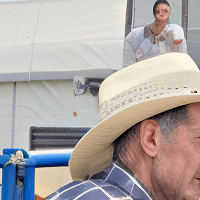}
      \vspace{-7mm}
      \caption*{ \fontsize{8.7pt}{8.7pt}\selectfont AddSR-s4}
    }  &
    \parbox[t]{0.135\textwidth}{
      \centering
      \includegraphics[width=\linewidth]{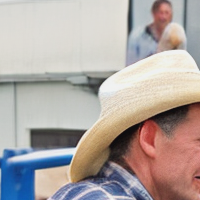}
      \vspace{-7mm}
      \caption*{ \fontsize{8.7pt}{8.7pt}\selectfont OSEDiff-s1}
    }  &
    \parbox[t]{0.135\textwidth}{
      \centering
      \includegraphics[width=\linewidth]{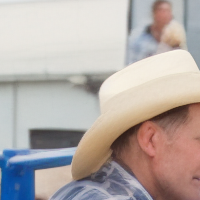}
      \vspace{-7mm}
      \caption*{ \fontsize{8.7pt}{8.7pt}\selectfont TSD-SR-s1}
    }  &
    \parbox[t]{0.135\textwidth}{
      \centering
      \includegraphics[width=\linewidth]{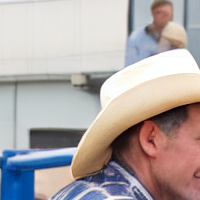}
      \vspace{-7mm}
      \caption*{ \fontsize{8.7pt}{8.7pt}\selectfont ICM-SR-s1}
    }  &
    \parbox[t]{0.135\textwidth}{
      \centering
      \includegraphics[width=\linewidth]{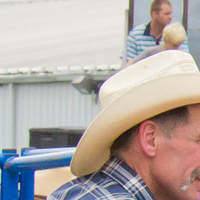}
      \vspace{-7mm}
      \caption*{ \fontsize{8.7pt}{8.7pt}\selectfont GT}
    }
    \vspace{-1mm}
  \end{tabular}
}
  \vspace{-2mm}
  \caption{Qualitative comparison of our method with various multi-step and one-step diffusion-based methods. ‘s’ denotes the number of network inferences in the method. Zoom in for better visualization.}
  \vspace{-2mm}
  \label{fig:qual_all}
\end{figure*}

\vspace{-2mm}
\paragraph{Evaluation details}
We evaluate the performance of our method on three benchmark datasets: DIV2K~\cite{agustsson2017ntire} validation dataset, RealSR~\cite{cai2019toward}, and DRealSR~\cite{wei2020component} provided
by StableSR~\cite{wang2024exploiting}.
For a thorough evaluation, we utilize a diverse set of both full-reference and no-reference image quality metrics.
For full-reference evaluation, we use PSNR and SSIM for fidelity, LPIPS~\cite{zhang2018unreasonable} and DISTS~\cite{ding2020image} for perceptual similarity, and FID~\cite{heusel2017gans} for distribution similarity.
For no-reference evaluation, we utilize NIQE~\cite{zhang2015feature}, MUSIQ~\cite{ke2021musiq}, MANIQA~\cite{yang2022maniqa}, and CLIPIQA~\cite{wang2023exploring}, TOPIQ~\cite{chen2024topiq}, and LIQE~\cite{zhang2023blind}.

\subsection{Quantitative comparison}
\label{subsec:exp_quant}

\cref{tab:main_comp} presents a quantitative comparison of the proposed method, ICM-SR, with the state-of-the-art one-step super-resolution models on both synthetic and real-world benchmarks.
For a fair comparison, we reproduce OSEDiff~\cite{wu2024one} and TSD-SR~\cite{dong2025tsd} using the same backbone~(Stable Diffusion v1-4) and training dataset.

As evident in \cref{tab:main_comp}, ICM-SR consistently excels across all benchmarks, paricularly for both full-reference perceptual metrics~(\eg, LPIPS, DISTS, FID) and no-reference quality metrics~(\eg, MUSIQ, MANIQA).
ICM-SR's strong performance in both metric categories highlights its superior ability to harmonize fidelity with perceptual realism.
In contrast, while TSD-SR also improves upon OSEDiff in full-reference metrics, it struggles with no-reference metrics.
This suggests that TSD-SR's approach, while enhancing fidelity, may fail to fully leverage the generative prior's capacity for producing aesthetically pleasing images.

The strong performance of ICM-SR extends to the challenging real-world datasets, such as DrealSR and RealSR, where low-quality~(LQ) images are sourced from practical scenarios.
This robust performance on practical, unkonwn degradations confirms the effectiveness of image-conditioned manifold regularization.
By providing a more stable and task-aligned regularization signal, ICM-SR successfully generates faithful and visually appealing super-resolution results.
For a comparison to multi-step methods, please refer to the supplementary material.

\begin{table*}[t]
	\centering
	\caption{
    Ablation studies of ICM-SR on the choice of structural image condition $\F_c$.
    The best and second best results of each metric are highlighted in \textcolor{red}{\textbf{red}} and \textcolor{blue}{blue}, respectively.
  }
  % !TEX root = ./../main.tex

\renewcommand{\arraystretch}{1} % Increase row spacing
\setlength\tabcolsep{4.5pt}
\scalebox{0.85}{
  \begin{tabular}{cl|ccc|cccccc|cc}
  \toprule
    \multirow{2}{*}{Datasets} & \multirow{2}{*}{Methods} & \multicolumn{3}{c|}{Perceptual w/ ref.} & \multicolumn{6}{c|}{ Perceptual w/o ref.} & \multicolumn{2}{c}{Fidelity w/ ref.} \\
    {} & {}           & {LPIPS$\downarrow$}      & {DISTS$\downarrow$}      & {FID$\downarrow$}        & {NIQE$\downarrow$}       & {MUSIQ$\uparrow$}       & {MANIQ$\uparrow$}        & {CLIPIQ$\uparrow$}       & {TOPIQ$\uparrow$}        & {LIQE$\uparrow$}         & {PSNR$\uparrow$}        & {SSIM$\uparrow$}         \\ \midrule
    % {Datasets} & {Methods}           & {LPIPS$\downarrow$}       & {DISTS$\downarrow$}       & {FID$\downarrow$}        & {NIQE$\downarrow$}        & {MUSIQ$\uparrow$}        & {MANIQ$\uparrow$}         & {CLIPIQA$\uparrow$}       & {TOPIQ$\uparrow$}        & {LIQE$\uparrow$}         & {PSNR$\uparrow$}         & {SSIM$\uparrow$}          \\ \hline
    \multirow{5}{*}{DIV2K-Val}
               & OSEDiff${}^\dagger$ & 0.2847                    & 0.1905                    & 26.15                    & {4.4918}                  & {67.73}                  & {0.6081}                  & 0.6394                    & 0.6014                   & {4.1704}                 & 23.40                    & 0.6160                    \\
               & ICM-SR-lq           & 0.3042                    & 0.1977                    & 28.32                    & 4.7224                    & 67.47                    & 0.6135                    & 0.6446                    & 0.5794                   & 4.1874                   & 22.72                    & 0.6044                    \\
               & ICM-SR-color        & 0.2849                    & 0.1877                    & 25.49                    & \textcolor{red}{\textbf{4.4361}} & \textcolor{red}{\textbf{68.64}} & \textcolor{red}{\textbf{0.6192}} & \textcolor{red}{\textbf{0.6520}} & \textcolor{red}{\textbf{0.6235}}  & \textcolor{red}{\textbf{4.2722}}  & 23.43                    & 0.6158                    \\
               & ICM-SR-canny        & \textcolor{red}{\textbf{0.2791}} & \textcolor{red}{\textbf{0.1856}} & \textcolor{red}{\textbf{24.66}} & 4.4630                    & 67.84                    & 0.6143                    & \textcolor{blue}{0.6457}  & 0.6089                   & 4.1967                   & \textcolor{blue}{23.66}  & \textcolor{red}{\textbf{0.6180}} \\
               & ICM-SR-multi & \textcolor{blue}{0.2799}  & \textcolor{blue}{0.1861}  & \textcolor{blue}{24.72}  & \textcolor{blue}{4.4411}  & \textcolor{blue}{68.00}  & \textcolor{blue}{0.6169}  & 0.6440                    & \textcolor{blue}{0.6138} & \textcolor{blue}{4.2094} & \textcolor{red}{\textbf{23.77}} & \textcolor{blue}{0.6173}  \\ \bottomrule
  \end{tabular}%
}

  \label{tab:ablation} 
\end{table*}

\subsection{Qualitative comparison}
\label{subsec:exp_qual}

\cref{fig:qual_osediff} provides a qualitative comparison of ICM-SR, OSEDiff, and TSD-SR, highlighting ICM-SR's ability to produce highly expressive textures that are faithful to the ground truth.
This qualitative superiority aligns with our strong performance in perceptual quality measures, as detailed in \cref{tab:main_comp}.
In the first super-resolution example in \cref{fig:qual_osediff}, ICM-SR excels at recovering fine details from the LQ input, clearly reconstructing the contour of the eyes while OSEDiff and TSD-SR fail.
Notably, ICM-SR's reconstruction inherits naturalness from pretrained diffusion models, enabling it to surpass the quality of the ground-truth image, which containing some noise.
In the second example in \cref{fig:qual_osediff}, ICM-SR demonstrates its superior ability to restore complex textures. 
While OSEDiff produces a smoothed-out, blurred texture that loses the intricate details of the wood grain, ICM-SR accurately recovers the deep grooves and unique patterns present in the ground truth. 
This results in an output that is far more faithful to the original material and perceptually more realistic.
Additionally, \cref{fig:qual_all} provides a broader qualitative comparison against various state-of-the-art diffusion-based SR methods, spanning both multi-step and one-step models. 
Across all examples, ICM-SR yields results that are simultaneously the sharpest and most consistent with the ground truth.
For example, our method successfully recovers the clear edges of the leaves and branches in the first example, and the fine details of the ear in the second example.
In contrast, other methods tend to either lose fine details to blurring or generate textures that deviate from the original.
For more qualitative results, please see supplementary material.

\subsection{Ablation study}
\label{subsec:ablation}

\cref{tab:ablation} presents a quantitative comparison of the proposed method, ICM-SR, with various structural image information. 
As discussed in \cref{subsec:icm}, directly conditioning the manifold with LQ images shows poor performance.
Interestingly, we observe that conditioning on the colormap, which preserves broad color distributions, results in higher scores for no-reference perceptual metrics. 
Conversely, conditioning on canny edges, which explicitly defines structural outlines, leads to better performance on full-reference metrics that reward structural fidelity.
Notably, a combination of both the colormap and canny edges proves the most effective, outperforming OSEDiff across all metrics and datasets.

\vspace{2mm}

\subsection{Analysis on image conditioned score}
\label{subsec:analysis}

\cref{fig:practical} shows the impact of image-conditioning on score estimation across various diffusion timesteps.
At large timestep~($t=980$), denoised output without image-conditioning struggle to retain image structure and produce oversaturated image with totally different content.
This indicates that the extracted generative prior is inappropriate for Real-ISR regularization.
In contrast, predictions conditioned on the structural image information~(colormap and Canny edges) effectively capture overall color tone and image structure, even at large timestep.
This indicates that image-conditioned manifold is task-aligned and it enhances score estimation accuracy.

% !TEX root = ./../main.tex

\vspace{-1mm}

\section{Conclusion and future work}
\label{sec:conclusion}

\vspace{-1mm}

We introduced Image-Conditioned Manifold~(ICM) regularization which addressed the fundamental problem of conceptual mismatch in one-step Real-ISR methods that rely on text-conditioned generative priors.
We identified that text-conditioning is often misaligned with the image fidelity goal of Real-ISR and na\"ive integration of dense image conditions may lead to numerical instability in the VSD loss.
The proposed ICM regularization resolves this dilemma by conditioning the target manifold on core structural information, colormaps and Canny edges.
This allows the teacher model to provide a stable and accurate prior with faithful colors and rich structural details.

Our comprehensive experiments demonstrated that ICM-SR achieved outstanding performance, particularly in perceptual quality metrics, highlighting its ability to generate visually appealing images that are faithful to LQ images.
While maintaining the efficiency of single-step inference, ICM-SR produced results with vivid textures and sharp details, often rivaling or surpassing the perceptual quality of multi-step diffusion methods.

While ICM-SR demonstrates strong performance, its reliance on large pretrained models results in a significant parameter count, and there remains opportunity for further refinement in restoring very fine details. 
Future work will explore model compression techniques and advanced conditioning strategies to further enhance the practical utility and performance in diverse real-world applications.

{
    \small
    \bibliographystyle{ieeenat_fullname}
    \bibliography{main}
}
\clearpage
\onecolumn
\appendix
% !TEX root = ./../main.tex

\section*{\Large{Appendix}}

\section{Quantitative comparison to multi-step super-resolution models}
\label{app:comp_multi}
Compared to multi-step methods~\cite{wang2024exploiting,lin2024diffbir, wu2024seesr, yu2024scaling, yang2024pixel, yue2023resshift, xie2024addsr}, ICM-SR outperforms in perceptual quality metrics with reference and delivers comparable results in no-reference perceptual metrics across all evaluated datasets~\cite{agustsson2017ntire, cai2019toward, wei2020component}. 
Given that ICM-SR operates as a one-step model, unlike the multi-step approaches, it achieves impressive perceptual quality with significantly lower computational overhead.

\begin{table*}[h]
    \centering
    \caption{
    Quantitative comparison with state-of-the-art methods on both synthetic and real-world benchmarks. `s' denotes the number of network inferences in the method.
    The best and second best results of each metric are highlighted in \textcolor{red}{\textbf{red}} and \textcolor{blue}{blue}, respectively.}
    % !TEX root = ./../appendix.tex

\renewcommand{\arraystretch}{1.1} % Increase row spacing
\setlength\tabcolsep{5pt}
\scalebox{0.9}{
  \begin{tabular}{cl|ccc|cccc|cc}
    \toprule
    \multirow{2}{*}{Datasets}  & \multirow{2}{*}{Methods} & \multicolumn{3}{c|}{Perceptual w/ ref.} & \multicolumn{4}{c|}{ Perceptual w/o ref.} & \multicolumn{2}{c}{Fidelity w/ ref.}                                                                                                                                                                                                                 \\
    {}                         & {}                       & {LPIPS$\downarrow$}                     & {DISTS$\downarrow$}                       & {FID$\downarrow$}                    & {NIQE$\downarrow$}               & {MUSIQ$\uparrow$}               & {MANIQ$\uparrow$}                & {CLIPIQ$\uparrow$}               & {PSNR$\uparrow$}                & {SSIM$\uparrow$}                 \\ \midrule

    \multirow{8}{*}{DIV2K-Val} & StableSR-s200            & 0.3114                                  & 0.2048                                    & \textcolor{red}{\textbf{24.44}}      & \textcolor{blue}{4.7581}         & 65.92                           & 0.6190                           & 0.6771                           & 23.26                           & 0.5726                           \\
                               & DiffBIR-s50              & 0.3669                                  & 0.2209                                    & 32.70                                & 4.9903                           & \textcolor{red}{\textbf{69.87}} & \textcolor{red}{\textbf{0.6461}} & \textcolor{blue}{0.7299}         & 23.14                           & 0.5441                           \\
                               & SeeSR-s50                & 0.3194                                  & \textcolor{blue}{0.1966}                  & 25.82                                & 4.7927                           & 68.42                           & 0.6219                           & 0.6867                           & 23.73                           & 0.6056                           \\
                               & SUPIR-s50                & 0.3919                                  & 0.2312                                    & 31.40                                & 5.6767                           & 63.86                           & 0.5903                           & 0.7146                           & 22.13                           & 0.5279                           \\
                               & PASD-s20                 & 0.3779                                  & 0.2305                                    & 39.12                                & 4.8587                           & 67.36                           & 0.6121                           & 0.6327                           & \textcolor{blue}{24.00}         & 0.6041                           \\
                               & ResShift-s15             & \textcolor{blue}{0.3077}                & 0.2136                                    & 30.79                                & 6.9152                           & 58.89                           & 0.5283                           & 0.5717                           & \textcolor{red}{\textbf{24.59}} & \textcolor{red}{\textbf{0.6232}} \\
                               & AddSR-s4                 & 0.3816                                  & 0.2340                                    & 34.71                                & 5.8441                           & \textcolor{blue}{69.18}         & \textcolor{blue}{0.6324}         & \textcolor{red}{\textbf{0.7532}} & 22.38                           & 0.5557                           \\
                               & ICM-SR-s1                & \textcolor{red}{\textbf{0.2799}}        & \textcolor{red}{\textbf{0.1861}}          & \textcolor{blue}{24.72}              & \textcolor{red}{\textbf{4.4411}} & 68.00                           & 0.6169                           & 0.6440                           & 23.77                           & \textcolor{blue}{0.6173}         \\ \hline
    \multirow{8}{*}{DrealSR}   & StableSR-s200            & 0.3284                                  & \textcolor{blue}{0.2269}                  & 148.95                               & 6.5239                           & 58.51                           & 0.5586                           & 0.6357                           & \textcolor{blue}{28.03}         & 0.7536                           \\
                               & DiffBIR-s50              & 0.4669                                  & 0.2882                                    & 180.52                               & \textcolor{red}{\textbf{6.3293}} & \textcolor{red}{\textbf{66.15}} & \textcolor{red}{\textbf{0.6230}} & \textcolor{blue}{0.7068}         & 25.91                           & 0.6245                           \\
                               & SeeSR-s50                & \textcolor{blue}{0.3142}                & 0.2299                                    & \textcolor{blue}{146.85}             & 6.4825                           & 64.74                           & 0.6005                           & 0.6895                           & \textcolor{red}{\textbf{28.14}} & \textcolor{blue}{0.7711}         \\
                               & SUPIR-s50                & 0.4243                                  & 0.2795                                    & 169.48                               & 7.3918                           & 58.79                           & 0.5471                           & 0.6749                           & 25.09                           & 0.6460                           \\
                               & PASD-s20                 & 0.3579                                  & 0.2524                                    & 171.03                               & 6.7661                           & 63.23                           & 0.5919                           & 0.6242                           & 27.79                           & 0.7495                           \\
                               & ResShift-s15             & 0.3870                                  & 0.2632                                    & 160.16                               & 8.6344                           & 51.23                           & 0.4644                           & 0.5399                           & 27.05                           & 0.7404                           \\
                               & AddSR-s4                 & 0.3709                                  & 0.2662                                    & 169.34                               & 7.9004                           & \textcolor{blue}{65.23}         & 0.6014                           & \textcolor{red}{\textbf{0.7153}} & 26.66                           & 0.7406                           \\
                               & ICM-SR-s1                & \textcolor{red}{\textbf{0.2871}}        & \textcolor{red}{\textbf{0.2142}}          & \textcolor{red}{\textbf{125.30}}     & \textcolor{blue}{6.4163}         & 65.96                           & \textcolor{blue}{0.6051}         & 0.6929                           & 26.85                           & \textcolor{red}{\textbf{0.7763}} \\ \hline
    \multirow{8}{*}{RealSR}    & StableSR-s200            & \textcolor{blue}{0.3002}                & \textcolor{blue}{0.2139}                  & 128.49                               & 5.8809                           & 65.88                           & 0.6249                           & 0.6234                           & 24.65                           & 0.7080                           \\
                               & DiffBIR-s50              & 0.3650                                  & 0.2399                                    & 130.67                               & 5.8335                           & 69.28                           & 0.6511                           & \textcolor{blue}{0.7051}         & 24.83                           & 0.6501                           \\
                               & SeeSR-s50                & 0.3004                                  & 0.2218                                    & \textcolor{blue}{125.09}             & \textcolor{red}{\textbf{5.3938}} & \textcolor{blue}{69.69}         & 0.6453                           & 0.6674                           & 25.20                           & 0.7215                           \\
                               & SUPIR-s50                & 0.3541                                  & 0.2488                                    & 130.38                               & 6.1099                           & 62.09                           & 0.5780                           & 0.6707                           & 23.65                           & 0.6620                           \\
                               & PASD-s20                 & 0.3144                                  & 0.2304                                    & 134.18                               & 5.7616                           & 68.33                           & 0.6323                           & 0.5783                           & \textcolor{red}{\textbf{25.68}} & 0.7273                           \\
                               & ResShift-s15             & 0.3279                                  & 0.2475                                    & 128.03                               & 8.0708                           & 56.88                           & 0.5100                           & 0.5362                           & \textcolor{blue}{25.66}         & \textcolor{red}{\textbf{0.7360}} \\
                               & AddSR-s4                 & 0.3820                                  & 0.2688                                    & 153.35                               & 6.4357                           & \textcolor{red}{\textbf{71.87}} & \textcolor{red}{\textbf{0.6767}} & \textcolor{red}{\textbf{0.7306}} & 22.53                           & 0.6452                           \\
                               & ICM-SR-s1                & \textcolor{red}{\textbf{0.2611}}        & \textcolor{red}{\textbf{0.2009}}          & \textcolor{red}{\textbf{108.74}}     & \textcolor{blue}{5.5842}         & 68.59                           & \textcolor{blue}{0.6511}         & 0.6360                           & 24.99                           & \textcolor{blue}{0.7309}         \\ \bottomrule
  \end{tabular}%
}

    \label{tab:comp_multi}
\end{table*}

\clearpage
\section{Ablation studies}
\label{app:ablation}

\subsection{Adapter scale}
\label{app:adapter_scale}
\begin{table*}[h]
	\centering
	\caption{
        Ablation study on the scale of the T2I Adapter. 
        We investigate the effect of varying the adapter scale values ($\{0.0, 0.5, 1.0\}$) on the super-resolution performance evaluated on the DIV2K validation dataset.
  }
  % !TEX root = ./../appendix.tex

\renewcommand{\arraystretch}{1} % Increase row spacing
\setlength\tabcolsep{4.5pt}
\scalebox{0.85}{
  \begin{tabular}{cc|ccc|cccccc|cc}
    \toprule
    \multirow{2}{*}{Methods} & \multirow{2}{*}{Scale} & \multicolumn{3}{c|}{Perceptual w/ ref.} & \multicolumn{6}{c|}{Perceptual w/o ref.} & \multicolumn{2}{c}{Fidelity w/ ref.}                                                                                                                                                                \\
    {}                       & {}                     & {LPIPS$\downarrow$}                     & {DISTS$\downarrow$}                      & {FID$\downarrow$}                    & {NIQE$\downarrow$} & {MUSIQ$\uparrow$} & {MANIQ$\uparrow$} & {CLIPIQ$\uparrow$} & {TOPIQ$\uparrow$} & {LIQE$\uparrow$} & {PSNR$\uparrow$} & {SSIM$\uparrow$} \\ \midrule
    OSEDiff                  & 0                      & 0.2847                   & 0.1905                   & 26.15                    & 4.4918 & 67.73 & 0.6081 & 0.6394                   & 0.6014                   & 4.1704 & 23.40                   & 0.6160                   \\ \midrule
    \multirow{2}{*}{ICM-SR}
                            %  & 0.2                    & 0.2800                                  & \textbf{0.1858}                          & 24.89                                & 4.4806             & 67.75             & 0.6146            & 0.6468             & 0.6116            & 4.2012           & 23.51            & 0.6174           \\
                             & 0.5                    & 0.2804                                  & 0.1862                                   & 25.09                                & 4.5090             & 67.92             & \textbf{0.6178}   & \textbf{0.6478}    & 0.6110            & \textbf{4.2117}  & 23.68            & \textbf{0.6193}  \\
                             & 1.0 (Ours)             & \textbf{0.2799}                         & \textbf{0.1861}                                   & \textbf{24.72}                       & \textbf{4.4411}    & \textbf{68.00}    & 0.6169            & 0.6440             & \textbf{0.6138}   & 4.2094           & \textbf{23.77}   & 0.6173           \\ \bottomrule
  \end{tabular}
}

  \label{tab:ablation_scale} 
\end{table*}  

To verify the impact of the structural guidance strength, we conduct an ablation study on the scale of the T2I-Adapter. 
We evaluate the super-resolution performance on the DIV2K validation dataset by varying the adapter scale values within $\{0.0, 0.5, 1.0\}$.
Note that setting the scale to $0.0$ is equivalent to the baseline method, OSEDiff~\cite{wu2024one}, which relies solely on text conditioning.
As presented in \cref{tab:ablation_scale}, introducing the core structural information~($\F_c$) via the T2I-Adapter significantly improves performance across all metrics compared to the baseline.
However, we observe that the performance benefits saturate above a certain threshold (e.g., 0.5) and, we adopt $1.0$ as the default setting for our main experiments.

\subsection{Updating strategy for the auxiliary diffusion model, $\epsilon_\psi$}
\label{app:update}

\begin{table*}[h]
	\centering
	\caption{
        Ablation study on the training strategy for $\epsilon_\psi$.
        The results are evaluated on the DIV2K validation dataset.
  }
  % !TEX root = ./../appendix.tex

\renewcommand{\arraystretch}{1} % Increase row spacing
\setlength\tabcolsep{4.5pt}
\scalebox{0.85}{
  \begin{tabular}{cc|ccc|cccccc|cc}
    \toprule
    \multirow{2}{*}{Methods} & \multirow{2}{*}{T2I-Adapter} & \multicolumn{3}{c|}{Perceptual w/ ref.} & \multicolumn{6}{c|}{Perceptual w/o ref.} & \multicolumn{2}{c}{Fidelity w/ ref.}                                                                                                                                                                \\
    {}                       & {}                      & {LPIPS$\downarrow$}                     & {DISTS$\downarrow$}                      & {FID$\downarrow$}                    & {NIQE$\downarrow$} & {MUSIQ$\uparrow$} & {MANIQ$\uparrow$} & {CLIPIQ$\uparrow$} & {TOPIQ$\uparrow$} & {LIQE$\uparrow$} & {PSNR$\uparrow$} & {SSIM$\uparrow$} \\ \midrule
    \multirow{2}{*}{ICM-SR}
                             & \cmark                  & \textbf{0.2785}                         & \textbf{0.1852}                          & \textbf{24.18}                       & 4.5000             & 67.15             & 0.6126            & 0.6307             & 0.5991            & 4.1165           & \textbf{23.96}   & \textbf{0.6202}  \\
                             & \xmark~ (Ours)          & 0.2799                                  & 0.1861                                   & 24.72                                & \textbf{4.4411}    & \textbf{68.00}    & \textbf{0.6169}   & \textbf{0.6440}    & \textbf{0.6138}   & \textbf{4.2094}  & 23.77            & 0.6173           \\ \bottomrule
  \end{tabular}
}
  \label{tab:ablation_update} 
\end{table*}

We conduct an ablation study on the training strategy for the auxiliary diffusion model~($\epsilon_\psi$), specifically investigating the effect of incorporating core structural information.
Formally, the objective of $\epsilon_\psi$ with T2I-Adapter is 
\begin{align*}
  ||\epsilon_\psi(\hat{\z}_t;t,\cc_t, A_\eta(\F_c)) - \epsilon||_2^2,
\end{align*}
which learns the conditional distribution of $G_\theta(\x_L)|\F_c$.
On the other hand, the objective of $\epsilon_\psi$ trained without T2I-Adapter~(our choice) is
\begin{align*}
  ||\epsilon_\psi(\hat{\z}_t;t,\cc_t) - \epsilon||_2^2,
\end{align*}
which learns the distribution of generator's output conditioned only on the text, \ie, $G_\theta(\x_L)s$.

As shown in \cref{tab:ablation_update}, the strategy incorporating the T2I-Adapter into the auxiliary model~($\checkmark$) yields slightly higher fidelity.
However, our chosen strategy without the adapter~(\xmark) demonstrates substantial gains in no-reference perceptual metrics with only marginal compromise in reference-based metrics.
In addition to this favorable performance trade-off, we consider the strategy without the adapter to be conceptually more appropriate. 
While ICM utilizes the T2I-Adapter to extract high-fidelity signal from pretrained diffusion models, the auxiliary model $\epsilon_\psi$ is tasked with learning the distribution of the generated output itself.
Providing the ground-truth structural condition ($F_c$) to $\epsilon_\psi$ makes the learning task trivial, potentially causing the model to over-rely on the condition instead of accurately capturing the distribution of the generated images.
Therefore, considering both the significant perceptual gains and the conceptual alignment, we adopt the strategy without the T2I-Adapter for $\epsilon_\psi$.

%%%%%%%%%%%%%%%%%%%%%%%%%%%%%%%%%%%%%%%%%%%%%%%%

% \input{tables/supple_sd1.4.tex} 
\clearpage
\section{Qualitative comparisons}
\label{app:qual}

In this section, we provide more qualitative examples in \cref{fig:qual_aaai_sup} on DIV2K~\citep{agustsson2017ntire} validation dataset, compared with other methods~\citep{wang2024exploiting, lin2024diffbir, wu2024seesr, yue2023resshift, xie2024addsr, dong2025tsd, wu2024one}.
For various LQ images, our method, ICM-SR, produces visually appealing super-resolution results.

% !TEX root = ./../appendix.tex

\begin{figure*}[h]
  \centering

\scalebox{0.75}{
  \begin{tabular}{@{}c@{\hspace{0.3em}}c@{\hspace{0.3em}}c@{\hspace{0.3em}}c@{\hspace{0.3em}}c@{\hspace{0.3em}}c@{\hspace{0.3em}}c@{}}
    \multirow{2}{*}[5.93em]{\parbox[t]{0.2981\textwidth}{
        \centering
        \vspace{-3.17mm}
        \includegraphics[width=\linewidth]{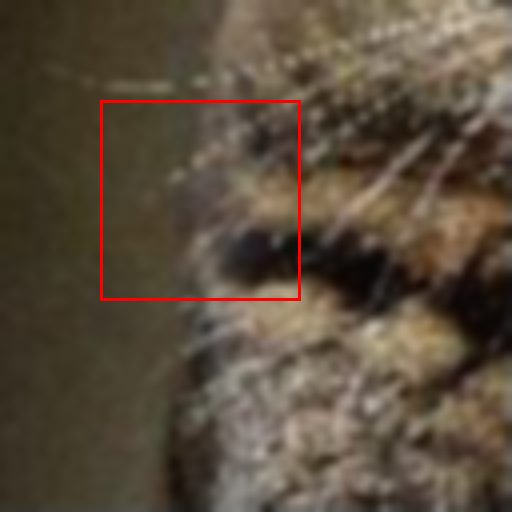}
        \vspace{-7mm}
        \caption*{ \fontsize{8.7pt}{8.7pt}\selectfont LQ}
    }} &
    \parbox[t]{0.135\textwidth}{
      \centering
      \includegraphics[width=\linewidth]{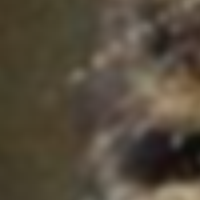}
      \vspace{-7mm}
      \caption*{ \fontsize{8.7pt}{8.7pt}\selectfont Zoomed LQ}
      \vspace{0mm}
    }  &
    \parbox[t]{0.135\textwidth}{
      \centering
      \includegraphics[width=\linewidth]{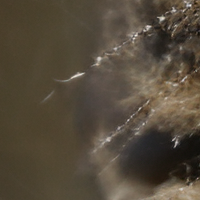}
      \vspace{-7mm}
      \caption*{ \fontsize{8.7pt}{8.7pt}\selectfont StableSR-200s}
    }  &
    \parbox[t]{0.135\textwidth}{
      \centering
      \includegraphics[width=\linewidth]{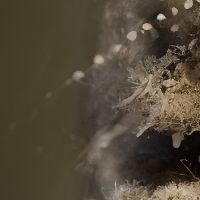}
      \vspace{-7mm}
      \caption*{ \fontsize{8.7pt}{8.7pt}\selectfont DiffBIR-50s}
    }  &
    \parbox[t]{0.135\textwidth}{
      \centering
      \includegraphics[width=\linewidth]{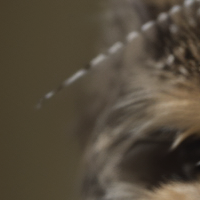}
      \vspace{-7mm}
      \caption*{ \fontsize{8.7pt}{8.7pt}\selectfont SeeSR-50s}
    }  &
    \parbox[t]{0.135\textwidth}{
      \centering
      \includegraphics[width=\linewidth]{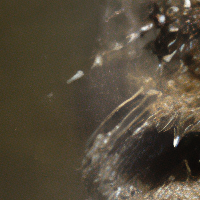}
      \vspace{-7mm}
      \caption*{ \fontsize{8.7pt}{8.7pt}\selectfont Resshift-15s}
    }    \\
    \vspace{2mm}
       &
    \parbox[t]{0.135\textwidth}{
      \centering
      \includegraphics[width=\linewidth]{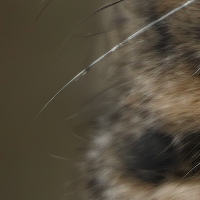}
      \vspace{-7mm}
      \caption*{ \fontsize{8.7pt}{8.7pt}\selectfont AddSR-4s}
    }  &
    \parbox[t]{0.135\textwidth}{
      \centering
      \includegraphics[width=\linewidth]{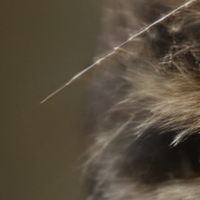} 
      \vspace{-7mm}
      \caption*{ \fontsize{8.7pt}{8.7pt}\selectfont OSEDiff-1s }
    } &
    \parbox[t]{0.135\textwidth}{
      \centering
      \includegraphics[width=\linewidth]{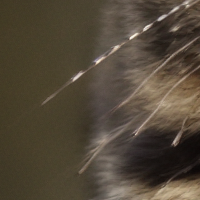} 
      \vspace{-7mm}
      \caption*{ \fontsize{8.7pt}{8.7pt}\selectfont TSD-SR-1s}
    }  &
    \parbox[t]{0.135\textwidth}{
      \centering
      \includegraphics[width=\linewidth]{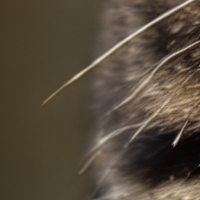} 
      \vspace{-7mm}
      \caption*{ \fontsize{8.7pt}{8.7pt}\selectfont ICM-SR-1s }
    } &
    \parbox[t]{0.135\textwidth}{
      \centering
      \includegraphics[width=\linewidth]{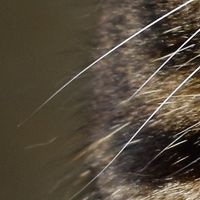} 
      \vspace{-7mm}
      \caption*{ \fontsize{8.7pt}{8.7pt}\selectfont GT }
    }
    \vspace{-2mm}
  \end{tabular}
}

  \scalebox{0.75}{
  \begin{tabular}{@{}c@{\hspace{0.3em}}c@{\hspace{0.3em}}c@{\hspace{0.3em}}c@{\hspace{0.3em}}c@{\hspace{0.3em}}c@{\hspace{0.3em}}c@{}}
    \multirow{2}{*}[5.93em]{\parbox[t]{0.2981\textwidth}{
        \centering
        \vspace{-3.17mm}
        \includegraphics[width=\linewidth]{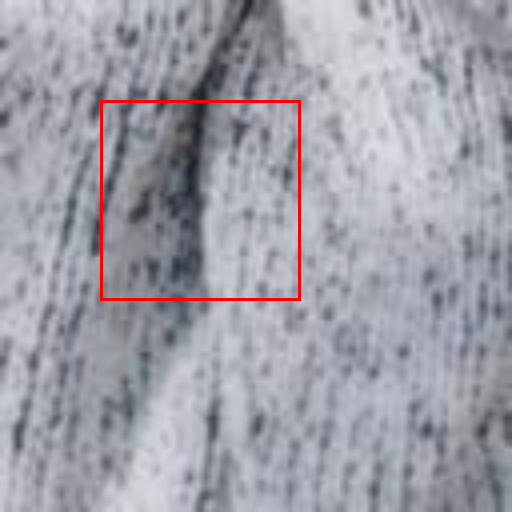}
        \vspace{-7mm}
        \caption*{ \fontsize{8.7pt}{8.7pt}\selectfont LQ}
    }} &
    \parbox[t]{0.135\textwidth}{
      \centering
      \includegraphics[width=\linewidth]{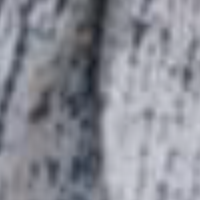}
      \vspace{-7mm}
      \caption*{ \fontsize{8.7pt}{8.7pt}\selectfont Zoomed LQ}
      \vspace{0mm}
    }  &
    \parbox[t]{0.135\textwidth}{
      \centering
      \includegraphics[width=\linewidth]{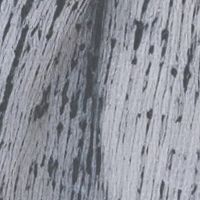}
      \vspace{-7mm}
      \caption*{ \fontsize{8.7pt}{8.7pt}\selectfont StableSR-200s}
    }  &
    \parbox[t]{0.135\textwidth}{
      \centering
      \includegraphics[width=\linewidth]{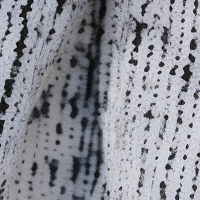}
      \vspace{-7mm}
      \caption*{ \fontsize{8.7pt}{8.7pt}\selectfont DiffBIR-50s}
    }  &
    \parbox[t]{0.135\textwidth}{
      \centering
      \includegraphics[width=\linewidth]{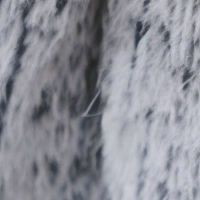}
      \vspace{-7mm}
      \caption*{ \fontsize{8.7pt}{8.7pt}\selectfont SeeSR-50s}
    }  &
    \parbox[t]{0.135\textwidth}{
      \centering
      \includegraphics[width=\linewidth]{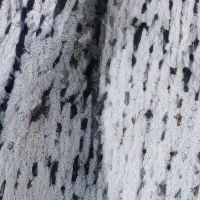}
      \vspace{-7mm}
      \caption*{ \fontsize{8.7pt}{8.7pt}\selectfont Resshift-15s}
    }    \\
    \vspace{2mm}
       &
    \parbox[t]{0.135\textwidth}{
      \centering
      \includegraphics[width=\linewidth]{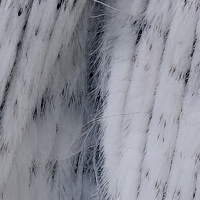}
      \vspace{-7mm}
      \caption*{ \fontsize{8.7pt}{8.7pt}\selectfont AddSR-4s}
    }  &
    \parbox[t]{0.135\textwidth}{
      \centering
      \includegraphics[width=\linewidth]{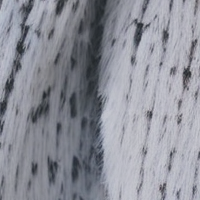} 
      \vspace{-7mm}
      \caption*{ \fontsize{8.7pt}{8.7pt}\selectfont OSEDiff-1s }
    } &
    \parbox[t]{0.135\textwidth}{
      \centering
      \includegraphics[width=\linewidth]{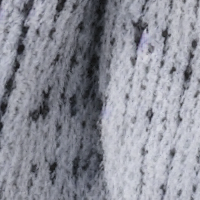} 
      \vspace{-7mm}
      \caption*{ \fontsize{8.7pt}{8.7pt}\selectfont TSD-SR-1s}
    }  &
    \parbox[t]{0.135\textwidth}{
      \centering
      \includegraphics[width=\linewidth]{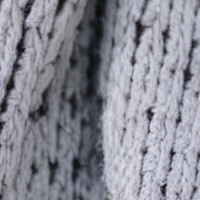} 
      \vspace{-7mm}
      \caption*{ \fontsize{8.7pt}{8.7pt}\selectfont ICM-SR-1s }
    } &
    \parbox[t]{0.135\textwidth}{
      \centering
      \includegraphics[width=\linewidth]{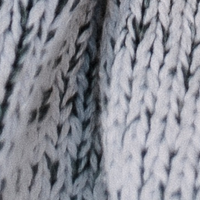} 
      \vspace{-7mm}
      \caption*{ \fontsize{8.7pt}{8.7pt}\selectfont GT }
    }
    \vspace{-2mm}
  \end{tabular}
}
  \scalebox{0.75}{
    \begin{tabular}{@{}c@{\hspace{0.3em}}c@{\hspace{0.3em}}c@{\hspace{0.3em}}c@{\hspace{0.3em}}c@{\hspace{0.3em}}c@{\hspace{0.3em}}c@{}}
      \multirow{2}{*}[5.93em]{\parbox[t]{0.2981\textwidth}{
          \centering
          \vspace{-3.17mm}
          \includegraphics[width=\linewidth]{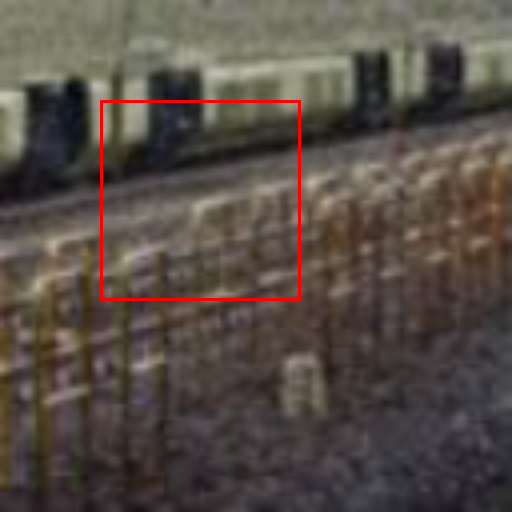}
          \vspace{-7mm}
          \caption*{ \fontsize{8.7pt}{8.7pt}\selectfont LQ}
      }} &
      \parbox[t]{0.135\textwidth}{
        \centering
        \includegraphics[width=\linewidth]{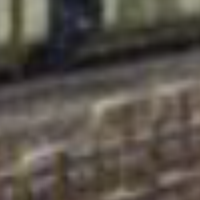}
        \vspace{-7mm}
        \caption*{ \fontsize{8.7pt}{8.7pt}\selectfont Zoomed LQ}
        \vspace{0mm}
      }  &
      \parbox[t]{0.135\textwidth}{
        \centering
        \includegraphics[width=\linewidth]{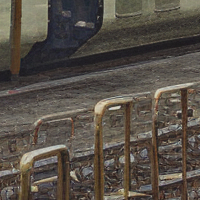}
        \vspace{-7mm}
        \caption*{ \fontsize{8.7pt}{8.7pt}\selectfont StableSR-200s}
      }  &
      \parbox[t]{0.135\textwidth}{
        \centering
        \includegraphics[width=\linewidth]{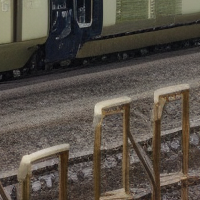}
        \vspace{-7mm}
        \caption*{ \fontsize{8.7pt}{8.7pt}\selectfont DiffBIR-50s}
      }  &
      \parbox[t]{0.135\textwidth}{
        \centering
        \includegraphics[width=\linewidth]{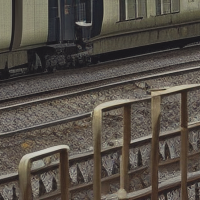}
        \vspace{-7mm}
        \caption*{ \fontsize{8.7pt}{8.7pt}\selectfont SeeSR-50s}
      }  &
      \parbox[t]{0.135\textwidth}{
        \centering
        \includegraphics[width=\linewidth]{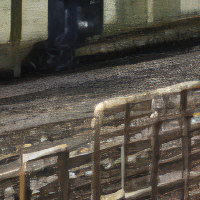}
        \vspace{-7mm}
        \caption*{ \fontsize{8.7pt}{8.7pt}\selectfont Resshift-15s}
      }    \\
      \vspace{2mm}
         &
      \parbox[t]{0.135\textwidth}{
        \centering
        \includegraphics[width=\linewidth]{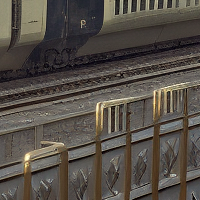}
        \vspace{-7mm}
        \caption*{ \fontsize{8.7pt}{8.7pt}\selectfont AddSR-4s}
      }  &
      \parbox[t]{0.135\textwidth}{
        \centering
        \includegraphics[width=\linewidth]{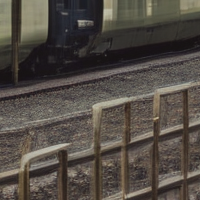} 
        \vspace{-7mm}
        \caption*{ \fontsize{8.7pt}{8.7pt}\selectfont OSEDiff-1s }
      } &
      \parbox[t]{0.135\textwidth}{
        \centering
        \includegraphics[width=\linewidth]{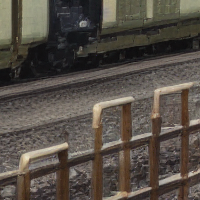} 
        \vspace{-7mm}
        \caption*{ \fontsize{8.7pt}{8.7pt}\selectfont TSD-SR-1s}
      }  &
      \parbox[t]{0.135\textwidth}{
        \centering
        \includegraphics[width=\linewidth]{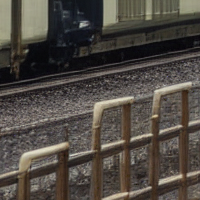} 
        \vspace{-7mm}
        \caption*{ \fontsize{8.7pt}{8.7pt}\selectfont ICM-SR-1s }
      } &
      \parbox[t]{0.135\textwidth}{
        \centering
        \includegraphics[width=\linewidth]{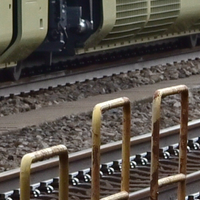} 
        \vspace{-7mm}
        \caption*{ \fontsize{8.7pt}{8.7pt}\selectfont GT }
      }
      \vspace{-2mm}
    \end{tabular}
  }

  \scalebox{0.75}{
  \begin{tabular}{@{}c@{\hspace{0.3em}}c@{\hspace{0.3em}}c@{\hspace{0.3em}}c@{\hspace{0.3em}}c@{\hspace{0.3em}}c@{\hspace{0.3em}}c@{}}
    \multirow{2}{*}[5.93em]{\parbox[t]{0.2981\textwidth}{
        \centering
        \vspace{-3.17mm}
        \includegraphics[width=\linewidth]{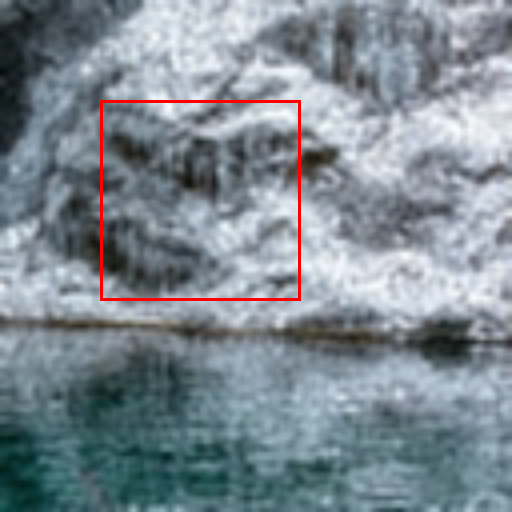}
        \vspace{-7mm}
        \caption*{ \fontsize{8.7pt}{8.7pt}\selectfont LQ}
    }} &
    \parbox[t]{0.135\textwidth}{
      \centering
      \includegraphics[width=\linewidth]{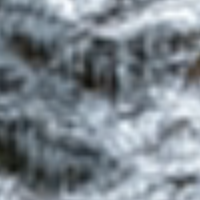}
      \vspace{-7mm}
      \caption*{ \fontsize{8.7pt}{8.7pt}\selectfont Zoomed LQ}
      \vspace{0mm}
    }  &
    \parbox[t]{0.135\textwidth}{
      \centering
      \includegraphics[width=\linewidth]{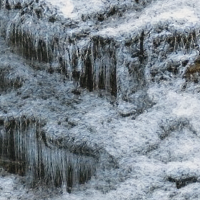}
      \vspace{-7mm}
      \caption*{ \fontsize{8.7pt}{8.7pt}\selectfont StableSR-200s}
    }  &
    \parbox[t]{0.135\textwidth}{
      \centering
      \includegraphics[width=\linewidth]{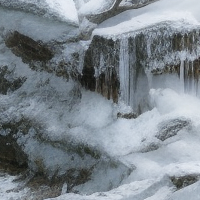}
      \vspace{-7mm}
      \caption*{ \fontsize{8.7pt}{8.7pt}\selectfont DiffBIR-50s}
    }  &
    \parbox[t]{0.135\textwidth}{
      \centering
      \includegraphics[width=\linewidth]{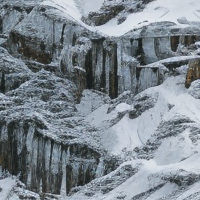}
      \vspace{-7mm}
      \caption*{ \fontsize{8.7pt}{8.7pt}\selectfont SeeSR-50s}
    }  &
    \parbox[t]{0.135\textwidth}{
      \centering
      \includegraphics[width=\linewidth]{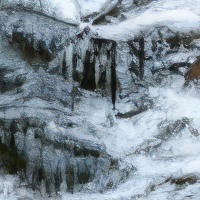}
      \vspace{-7mm}
      \caption*{ \fontsize{8.7pt}{8.7pt}\selectfont Resshift-15s}
    }    \\
    \vspace{2mm}
       &
    \parbox[t]{0.135\textwidth}{
      \centering
      \includegraphics[width=\linewidth]{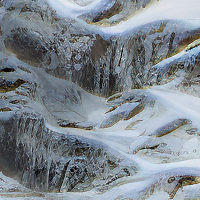}
      \vspace{-7mm}
      \caption*{ \fontsize{8.7pt}{8.7pt}\selectfont AddSR-4s}
    }  &
    \parbox[t]{0.135\textwidth}{
      \centering
      \includegraphics[width=\linewidth]{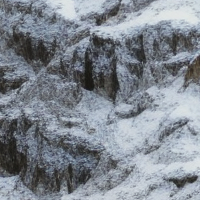} 
      \vspace{-7mm}
      \caption*{ \fontsize{8.7pt}{8.7pt}\selectfont OSEDiff-1s }
    } &
    \parbox[t]{0.135\textwidth}{
      \centering
      \includegraphics[width=\linewidth]{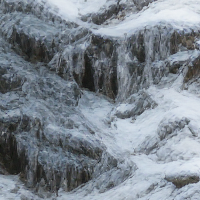} 
      \vspace{-7mm}
      \caption*{ \fontsize{8.7pt}{8.7pt}\selectfont TSD-SR-1s}
    }  &
    \parbox[t]{0.135\textwidth}{
      \centering
      \includegraphics[width=\linewidth]{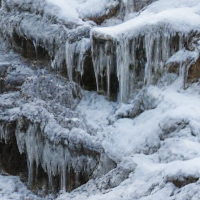} 
      \vspace{-7mm}
      \caption*{ \fontsize{8.7pt}{8.7pt}\selectfont ICM-SR-1s }
    } &
    \parbox[t]{0.135\textwidth}{
      \centering
      \includegraphics[width=\linewidth]{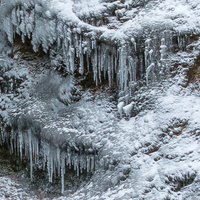} 
      \vspace{-7mm}
      \caption*{ \fontsize{8.7pt}{8.7pt}\selectfont GT }
    }
    \vspace{-2mm}
  \end{tabular}
}

  % \vspace{-3mm}
  \centering
  \caption{Qualitative results. Zoom in for better visualization.}
  \vspace{-2mm}
  \label{fig:qual_aaai_sup}
\end{figure*}

% !TEX root = ./../appendix.tex

\begin{figure*}[t]
  \centering

  \scalebox{0.75}{
  \begin{tabular}{@{}c@{\hspace{0.3em}}c@{\hspace{0.3em}}c@{\hspace{0.3em}}c@{\hspace{0.3em}}c@{\hspace{0.3em}}c@{\hspace{0.3em}}c@{}}
    \multirow{2}{*}[5.93em]{\parbox[t]{0.2981\textwidth}{
        \centering
        \vspace{-3.17mm}
        \includegraphics[width=\linewidth]{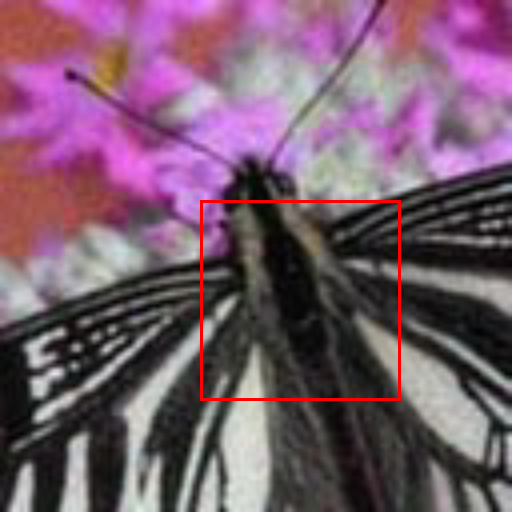}
        \vspace{-7mm}
        \caption*{ \fontsize{8.7pt}{8.7pt}\selectfont LQ}
    }} &
    \parbox[t]{0.135\textwidth}{
      \centering
      \includegraphics[width=\linewidth]{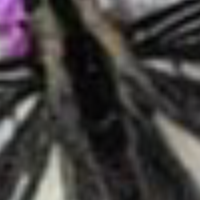}
      \vspace{-7mm}
      \caption*{ \fontsize{8.7pt}{8.7pt}\selectfont Zoomed LQ}
      \vspace{0mm}
    }  &
    \parbox[t]{0.135\textwidth}{
      \centering
      \includegraphics[width=\linewidth]{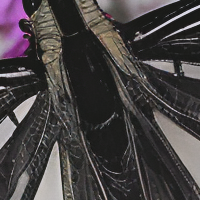}
      \vspace{-7mm}
      \caption*{ \fontsize{8.7pt}{8.7pt}\selectfont StableSR-200s}
    }  &
    \parbox[t]{0.135\textwidth}{
      \centering
      \includegraphics[width=\linewidth]{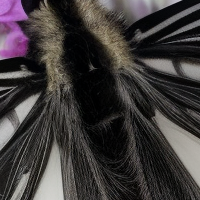}
      \vspace{-7mm}
      \caption*{ \fontsize{8.7pt}{8.7pt}\selectfont DiffBIR-50s}
    }  &
    \parbox[t]{0.135\textwidth}{
      \centering
      \includegraphics[width=\linewidth]{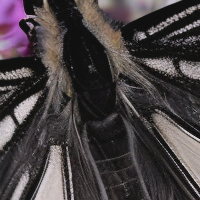}
      \vspace{-7mm}
      \caption*{ \fontsize{8.7pt}{8.7pt}\selectfont SeeSR-50s}
    }  &
    \parbox[t]{0.135\textwidth}{
      \centering
      \includegraphics[width=\linewidth]{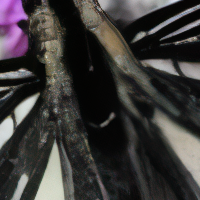}
      \vspace{-7mm}
      \caption*{ \fontsize{8.7pt}{8.7pt}\selectfont Resshift-15s}
    }    \\
    \vspace{2mm}
       &
    \parbox[t]{0.135\textwidth}{
      \centering
      \includegraphics[width=\linewidth]{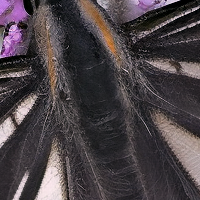}
      \vspace{-7mm}
      \caption*{ \fontsize{8.7pt}{8.7pt}\selectfont AddSR-4s}
    }  &
    \parbox[t]{0.135\textwidth}{
      \centering
      \includegraphics[width=\linewidth]{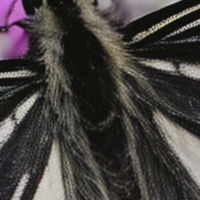} 
      \vspace{-7mm}
      \caption*{ \fontsize{8.7pt}{8.7pt}\selectfont OSEDiff-1s }
    } &
    \parbox[t]{0.135\textwidth}{
      \centering
      \includegraphics[width=\linewidth]{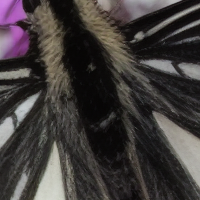} 
      \vspace{-7mm}
      \caption*{ \fontsize{8.7pt}{8.7pt}\selectfont TSD-SR-1s}
    }  &
    \parbox[t]{0.135\textwidth}{
      \centering
      \includegraphics[width=\linewidth]{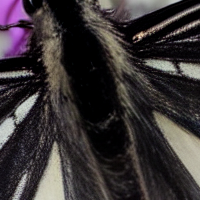} 
      \vspace{-7mm}
      \caption*{ \fontsize{8.7pt}{8.7pt}\selectfont ICM-SR-1s }
    } &
    \parbox[t]{0.135\textwidth}{
      \centering
      \includegraphics[width=\linewidth]{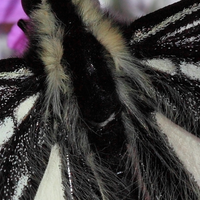} 
      \vspace{-7mm}
      \caption*{ \fontsize{8.7pt}{8.7pt}\selectfont GT }
    }
    \vspace{-2mm}
  \end{tabular}
}

\scalebox{0.75}{
  \begin{tabular}{@{}c@{\hspace{0.3em}}c@{\hspace{0.3em}}c@{\hspace{0.3em}}c@{\hspace{0.3em}}c@{\hspace{0.3em}}c@{\hspace{0.3em}}c@{}}
    \multirow{2}{*}[5.93em]{\parbox[t]{0.2981\textwidth}{
        \centering
        \vspace{-3.17mm}
        \includegraphics[width=\linewidth]{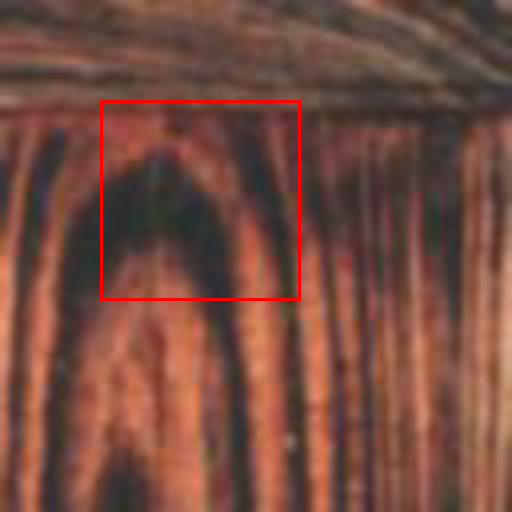}
        \vspace{-7mm}
        \caption*{ \fontsize{8.7pt}{8.7pt}\selectfont LQ}
    }} &
    \parbox[t]{0.135\textwidth}{
      \centering
      \includegraphics[width=\linewidth]{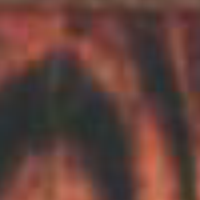}
      \vspace{-7mm}
      \caption*{ \fontsize{8.7pt}{8.7pt}\selectfont Zoomed LQ}
      \vspace{0mm}
    }  &
    \parbox[t]{0.135\textwidth}{
      \centering
      \includegraphics[width=\linewidth]{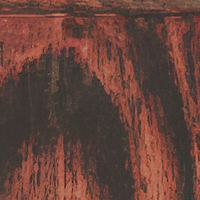}
      \vspace{-7mm}
      \caption*{ \fontsize{8.7pt}{8.7pt}\selectfont StableSR-200s}
    }  &
    \parbox[t]{0.135\textwidth}{
      \centering
      \includegraphics[width=\linewidth]{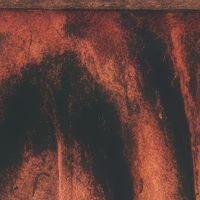}
      \vspace{-7mm}
      \caption*{ \fontsize{8.7pt}{8.7pt}\selectfont DiffBIR-50s}
    }  &
    \parbox[t]{0.135\textwidth}{
      \centering
      \includegraphics[width=\linewidth]{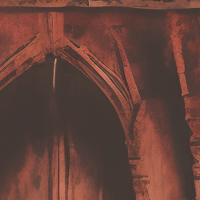}
      \vspace{-7mm}
      \caption*{ \fontsize{8.7pt}{8.7pt}\selectfont SeeSR-50s}
    }  &
    \parbox[t]{0.135\textwidth}{
      \centering
      \includegraphics[width=\linewidth]{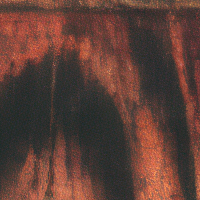}
      \vspace{-7mm}
      \caption*{ \fontsize{8.7pt}{8.7pt}\selectfont Resshift-15s}
    }    \\
    \vspace{2mm}
       &
    \parbox[t]{0.135\textwidth}{
      \centering
      \includegraphics[width=\linewidth]{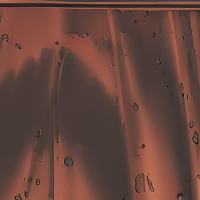}
      \vspace{-7mm}
      \caption*{ \fontsize{8.7pt}{8.7pt}\selectfont AddSR-4s}
    }  &
    \parbox[t]{0.135\textwidth}{
      \centering
      \includegraphics[width=\linewidth]{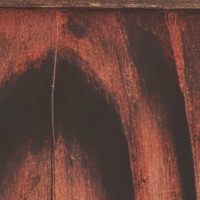} 
      \vspace{-7mm}
      \caption*{ \fontsize{8.7pt}{8.7pt}\selectfont OSEDiff-1s }
    } &
    \parbox[t]{0.135\textwidth}{
      \centering
      \includegraphics[width=\linewidth]{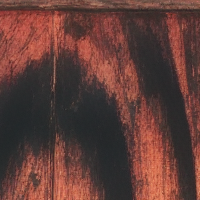} 
      \vspace{-7mm}
      \caption*{ \fontsize{8.7pt}{8.7pt}\selectfont TSD-SR-1s}
    }  &
    \parbox[t]{0.135\textwidth}{
      \centering
      \includegraphics[width=\linewidth]{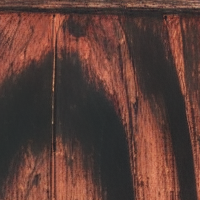} 
      \vspace{-7mm}
      \caption*{ \fontsize{8.7pt}{8.7pt}\selectfont ICM-SR-1s }
    } &
    \parbox[t]{0.135\textwidth}{
      \centering
      \includegraphics[width=\linewidth]{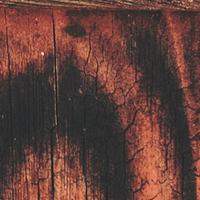} 
      \vspace{-7mm}
      \caption*{ \fontsize{8.7pt}{8.7pt}\selectfont GT }
    }
    \vspace{-2mm}
  \end{tabular}
}

\scalebox{0.75}{
  \begin{tabular}{@{}c@{\hspace{0.3em}}c@{\hspace{0.3em}}c@{\hspace{0.3em}}c@{\hspace{0.3em}}c@{\hspace{0.3em}}c@{\hspace{0.3em}}c@{}}
    \multirow{2}{*}[5.93em]{\parbox[t]{0.2981\textwidth}{
        \centering
        \vspace{-3.17mm}
        \includegraphics[width=\linewidth]{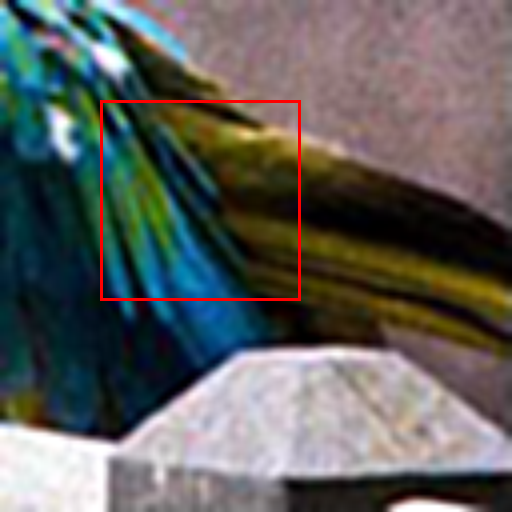}
        \vspace{-7mm}
        \caption*{ \fontsize{8.7pt}{8.7pt}\selectfont LQ}
    }} &
    \parbox[t]{0.135\textwidth}{
      \centering
      \includegraphics[width=\linewidth]{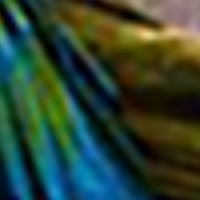}
      \vspace{-7mm}
      \caption*{ \fontsize{8.7pt}{8.7pt}\selectfont Zoomed LQ}
      \vspace{0mm}
    }  &
    \parbox[t]{0.135\textwidth}{
      \centering
      \includegraphics[width=\linewidth]{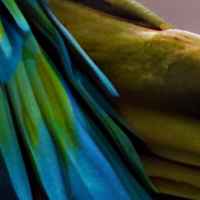}
      \vspace{-7mm}
      \caption*{ \fontsize{8.7pt}{8.7pt}\selectfont StableSR-200s}
    }  &
    \parbox[t]{0.135\textwidth}{
      \centering
      \includegraphics[width=\linewidth]{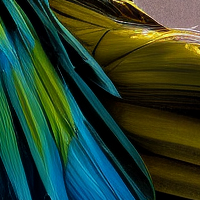}
      \vspace{-7mm}
      \caption*{ \fontsize{8.7pt}{8.7pt}\selectfont DiffBIR-50s}
    }  &
    \parbox[t]{0.135\textwidth}{
      \centering
      \includegraphics[width=\linewidth]{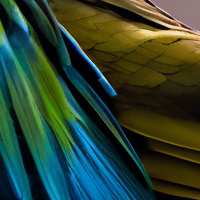}
      \vspace{-7mm}
      \caption*{ \fontsize{8.7pt}{8.7pt}\selectfont SeeSR-50s}
    }  &
    \parbox[t]{0.135\textwidth}{
      \centering
      \includegraphics[width=\linewidth]{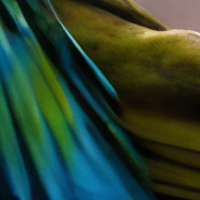}
      \vspace{-7mm}
      \caption*{ \fontsize{8.7pt}{8.7pt}\selectfont Resshift-15s}
    }    \\
    \vspace{2mm}
       &
    \parbox[t]{0.135\textwidth}{
      \centering
      \includegraphics[width=\linewidth]{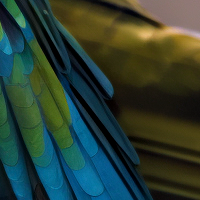}
      \vspace{-7mm}
      \caption*{ \fontsize{8.7pt}{8.7pt}\selectfont AddSR-4s}
    }  &
    \parbox[t]{0.135\textwidth}{
      \centering
      \includegraphics[width=\linewidth]{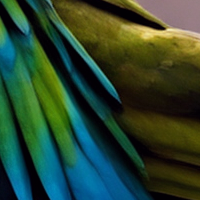} 
      \vspace{-7mm}
      \caption*{ \fontsize{8.7pt}{8.7pt}\selectfont OSEDiff-1s }
    } &
    \parbox[t]{0.135\textwidth}{
      \centering
      \includegraphics[width=\linewidth]{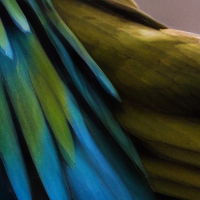} 
      \vspace{-7mm}
      \caption*{ \fontsize{8.7pt}{8.7pt}\selectfont TSD-SR-1s}
    }  &
    \parbox[t]{0.135\textwidth}{
      \centering
      \includegraphics[width=\linewidth]{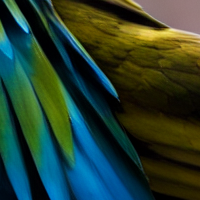} 
      \vspace{-7mm}
      \caption*{ \fontsize{8.7pt}{8.7pt}\selectfont ICM-SR-1s }
    } &
    \parbox[t]{0.135\textwidth}{
      \centering
      \includegraphics[width=\linewidth]{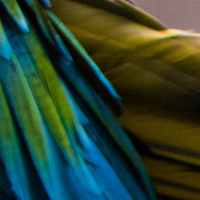} 
      \vspace{-7mm}
      \caption*{ \fontsize{8.7pt}{8.7pt}\selectfont GT }
    }
    \vspace{-2mm}
  \end{tabular}
}
\scalebox{0.75}{
  \begin{tabular}{@{}c@{\hspace{0.3em}}c@{\hspace{0.3em}}c@{\hspace{0.3em}}c@{\hspace{0.3em}}c@{\hspace{0.3em}}c@{\hspace{0.3em}}c@{}}
    \multirow{2}{*}[5.93em]{\parbox[t]{0.2981\textwidth}{
        \centering
        \vspace{-3.17mm}
        \includegraphics[width=\linewidth]{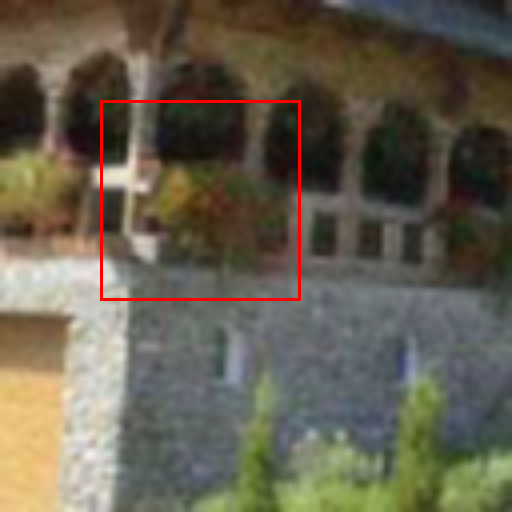}
        \vspace{-7mm}
        \caption*{ \fontsize{8.7pt}{8.7pt}\selectfont LQ}
    }} &
    \parbox[t]{0.135\textwidth}{
      \centering
      \includegraphics[width=\linewidth]{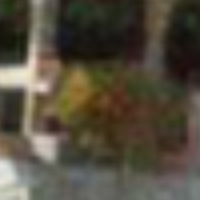}
      \vspace{-7mm}
      \caption*{ \fontsize{8.7pt}{8.7pt}\selectfont Zoomed LQ}
      \vspace{0mm}
    }  &
    \parbox[t]{0.135\textwidth}{
      \centering
      \includegraphics[width=\linewidth]{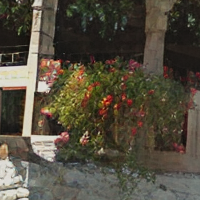}
      \vspace{-7mm}
      \caption*{ \fontsize{8.7pt}{8.7pt}\selectfont StableSR-200s}
    }  &
    \parbox[t]{0.135\textwidth}{
      \centering
      \includegraphics[width=\linewidth]{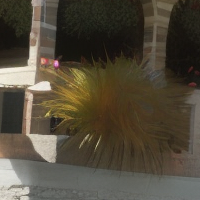}
      \vspace{-7mm}
      \caption*{ \fontsize{8.7pt}{8.7pt}\selectfont DiffBIR-50s}
    }  &
    \parbox[t]{0.135\textwidth}{
      \centering
      \includegraphics[width=\linewidth]{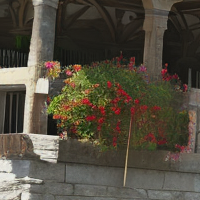}
      \vspace{-7mm}
      \caption*{ \fontsize{8.7pt}{8.7pt}\selectfont SeeSR-50s}
    }  &
    \parbox[t]{0.135\textwidth}{
      \centering
      \includegraphics[width=\linewidth]{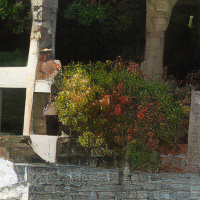}
      \vspace{-7mm}
      \caption*{ \fontsize{8.7pt}{8.7pt}\selectfont Resshift-15s}
    }    \\
    \vspace{2mm}
       &
    \parbox[t]{0.135\textwidth}{
      \centering
      \includegraphics[width=\linewidth]{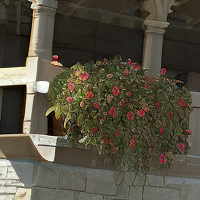}
      \vspace{-7mm}
      \caption*{ \fontsize{8.7pt}{8.7pt}\selectfont AddSR-4s}
    }  &
    \parbox[t]{0.135\textwidth}{
      \centering
      \includegraphics[width=\linewidth]{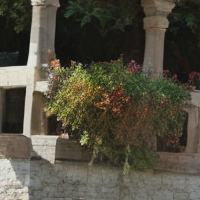} 
      \vspace{-7mm}
      \caption*{ \fontsize{8.7pt}{8.7pt}\selectfont OSEDiff-1s }
    } &
    \parbox[t]{0.135\textwidth}{
      \centering
      \includegraphics[width=\linewidth]{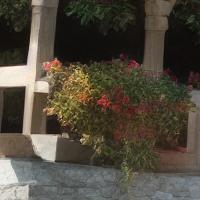} 
      \vspace{-7mm}
      \caption*{ \fontsize{8.7pt}{8.7pt}\selectfont TSD-SR-1s}
    }  &
    \parbox[t]{0.135\textwidth}{
      \centering
      \includegraphics[width=\linewidth]{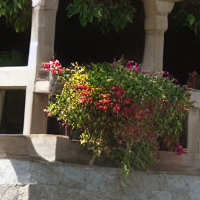} 
      \vspace{-7mm}
      \caption*{ \fontsize{8.7pt}{8.7pt}\selectfont ICM-SR-1s }
    } &
    \parbox[t]{0.135\textwidth}{
      \centering
      \includegraphics[width=\linewidth]{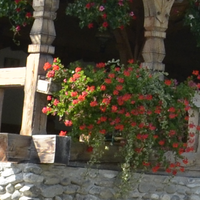} 
      \vspace{-7mm}
      \caption*{ \fontsize{8.7pt}{8.7pt}\selectfont GT }
    }
    \vspace{-2mm}
  \end{tabular}
}

  % \vspace{-3mm}
  \centering
  \caption{Qualitative results. Zoom in for better visualization.}
  \vspace{-2mm}
  \label{fig:qual_aaai_sup}
\end{figure*}

\end{document}